\pdfoutput=1 %
\documentclass[11pt]{article}

\usepackage[margin=1in]{geometry}
\usepackage[T1]{fontenc}
\usepackage{iftex}
\ifPDFTeX
  \usepackage[utf8]{inputenc}
\fi
\usepackage{lmodern}
\usepackage{microtype}
\usepackage{amsmath,amssymb}
\usepackage{graphicx}
\usepackage{booktabs,tabularx,longtable}
\usepackage{pdflscape}
\usepackage{xcolor}
\usepackage[numbers,sort&compress]{natbib}
\usepackage{hyperref}
\hypersetup{colorlinks=true,linkcolor=blue!50!black,citecolor=blue!50!black,urlcolor=blue!50!black,
  pdftitle={Parametric CAD Bench v3: Evaluating Agentic Parametric CAD Authoring from Text and Engineering Drawings},
  pdfauthor={Mei Chen, Yu Xia, Haipeng Mao}}
\newcommand{\bench}{Parametric CAD Bench v3}
\newcommand{\fcstd}{\texttt{FCStd}}
\newcommand{\code}[1]{\texttt{\detokenize{#1}}}

\title{\bench:\\Evaluating Agentic Parametric CAD Authoring\\from Text and Engineering Drawings}
\author{Mei Chen\\gNucleus AI, Inc.\\\texttt{mei.chen@gnucleus.ai}
  \and Yu Xia\\gNucleus AI, Inc.\\\texttt{peggy.xia@gnucleus.ai}
  \and Haipeng Mao\\gNucleus AI, Inc.\\\texttt{roc.mao@gnucleus.ai}}
\date{Results snapshot: September 24, 2026}

\begin{document}
\maketitle

\begin{abstract}
Producing an executable CAD script is not sufficient to establish that an
agent has satisfied an engineering design request. Evaluation must also
consider the resulting geometry, specified dimensions, and native CAD modeling
history. We present \bench, a 100-task benchmark comprising 30 text-to-CAD
creation tasks, 30 text-to-CAD create-and-edit tasks, and 40 tasks that
reconstruct mechanical parts from engineering drawings. Agents operate in a
pinned FreeCAD environment and submit native CAD artifacts to a separate
programmatic verifier. The primary metric is mean continuous task reward,
combining geometry similarity and specification consistency rather than binary
task success. A September 24, 2026 snapshot covers 13 model--agent systems;
Claude Code with Claude Opus 5.5 achieves the highest overall point estimate,
61.03\%, and an image-to-CAD mean reward of 84.66\%. These measurements
describe complete systems, not isolated model capabilities. We release
versioned tasks, evaluation configurations, and inspectable result artifacts.
\end{abstract}

\section{Introduction}
Computer-aided design (CAD) systems represent engineering intent as editable
digital models. In a parametric CAD system, the artifact contains more than
final geometry: a geometry kernel represents and manipulates curves, surfaces,
topology, and solids; constraint solvers enforce dimensional and geometric
relationships; and feature histories and dependency graphs record modeling
steps so that changes to upstream parameters or features propagate to
downstream geometry. These capabilities are fundamental to both professional
CAD systems such as Siemens NX, CATIA, Creo, and SolidWorks, and open-source
systems such as FreeCAD, which uses Open CASCADE Technology (OCCT) as its
geometry kernel. By contrast, CadQuery is a code-centric parametric CAD
library built on OCCT. Our benchmark instead requires native FreeCAD documents
with an application-managed feature tree and document dependencies, rather
than a CadQuery program or exported B-rep alone.

This distinction matters for evaluating CAD agents in real engineering
workflows, where parametric CAD systems underpin modern mechanical design and
manufacturing. An output can resemble the
requested part while containing incorrect dimensions, missing features,
broken dependencies, or a static shape in place of the required native
feature structure. Conversely, alternative feature constructions can satisfy
the same engineering requirements. Evaluating CAD agents therefore requires
more than testing whether generated code executes or whether the resulting
geometry is visually or geometrically similar.

Our central question is whether general-purpose AI agents can produce
geometrically correct, specification-consistent, parametric CAD artifacts
across three workflows.

The benchmark has evolved through three releases. We released Parametric CAD
Bench v1 in May 2026 to evaluate text-based creation of native parametric
FreeCAD models \citep{parametriccadbenchv1}. In September 2026, v2 retained
the 100-task creation format while tightening validation and evaluation
isolation: it upgraded the validator and FreeCAD runtime, moved held-out assets
into a separate verifier image, and standardized reward reporting and failure
handling \citep{parametriccadbenchv2}. V3 broadened the suite to include editing
and reconstruction from engineering drawings \citep{parametriccadbenchv3};
the September 24 update added evaluations of Claude Opus 5.5 with two agent
frameworks \citep{parametriccadbenchv3opus55}.

This paper focuses on v3; v1 and v2 provide historical context rather than
additional evaluation cohorts.

\bench\ evaluates three workflows: text-to-CAD creation, text-to-CAD
create-and-edit (two-stage creation and editing), and image-to-CAD
(reconstruction from engineering drawings); we refer to the first two as
creation and create-and-edit. Its
native FreeCAD output contract permits inspection beyond a rendered image,
while its grader combines geometric and task-specific specification checks.
These checks are proxies for the intended engineering requirements; they do
not by themselves certify manufacturing readiness or functional editability
under arbitrary future modifications.

The contribution of this work is threefold:
\begin{enumerate}
  \item A versioned 100-task evaluation suite spanning three CAD workflows with
        native parametric output requirements.
  \item An isolated execution and programmatic scoring protocol, with frozen
        configurations and per-task diagnostic artifacts.
  \item A dated evaluation of complete model--agent systems, exposing differences
        across workflows and observed failure modes.
\end{enumerate}

\section{Related Work}
\paragraph{CADGenBench: artifact-centered CAD evaluation.}
CADGenBench evaluates CAD generation and editing through tool-agnostic
STEP/BREP submissions \citep{cadgenbench,cadgenbenchsubmission}.
Its public leaderboard accepts a ZIP archive of candidate artifacts;
the hosted Hugging Face Space runs evaluation, and maintainers separately
review methodology evidence before marking a submission validated
\citep{cadgenbench,cadgenbenchvalidation}.
This is an artifact-submission protocol rather than a requirement to run
a submitted agent inside a common task environment. However, it is not
restricted to single-shot generation: its reference baseline is iterative,
and evaluation tools are also available locally
\citep{cadgenbench,cadgenbenchsubmission}.
Our emphasis is instead on evaluating complete agent executions in a
versioned FreeCAD environment and requiring native parametric documents.

\paragraph{CADBench: multimodal CAD-program reconstruction.}
MIT's CADBench evaluates reconstruction of executable, editable CAD programs
from meshes and rendered images \citep{mitcadbench}.
Its released evaluation pipeline currently accepts generated programs as
CadQuery Python code; supporting other CAD programming languages would
require adapting the pipeline \citep{mitcadbenchcode}.
It contains 18,000 samples across six benchmark families and five input
modalities, with complexity-stratified STEP-based splits and diversity-aware
sampling. Its six metrics cover volumetric IoU, surface IoU, Chamfer distance,
valid-shape rate, token count, and operation count.
Like \bench, it evaluates executable CAD geometry rather than rendered
appearance alone. However, its reported general-purpose VLM evaluation uses
direct generation without tool use, self-correction, or multi-turn refinement;
we evaluate complete interactive agent workflows in FreeCAD.
CADBench emphasizes reconstruction from 3D observations and object renders,
whereas our suite combines text-based creation, create-and-edit, and
reconstruction from dimensioned engineering drawings.
Its paper also distinguishes code editability from recovering the original
feature tree, sketch constraints, or design intent, which it does not require.
Our tasks instead require native FreeCAD documents with task-specific
parametric structure and specification checks, without requiring exact
reproduction of the reference construction history.
The benchmarks are complementary: CADBench provides broad reconstruction
and modality-shift diagnostics, while our smaller suite studies end-to-end
CAD task completion by model--agent systems.

\paragraph{BenchCAD: programmatic CAD over CadQuery.}
BenchCAD 1.0 contains 17,900 execution-verified CadQuery programs across
106 industrial part families, with image-to-code, instruction-guided code
editing, and visual/code question-answering tasks \citep{benchcad}.
Its generation evaluation compares the executed geometry with a reference;
the current website also reports agentic runs with rendering, measurement,
and iterative tool use \citep{benchcadwebsite}.
BenchCAD 2.0, released on October 12, 2026, after our September 24 snapshot,
adds part and assembly construction from dimensioned engineering drawings,
reverse engineering from rendered views, and agentic evaluation in an offline
sandbox, all scored against reference geometry \citep{benchcadtwo}. Its
drawing-based part tasks are the closest counterpart to our image-to-CAD
workflow.
The distinction from \bench\ is therefore not whether agents can use tools,
but the modeling interface and required output representation.
CadQuery is a high-level parametric CAD library built on Python bindings
to Open CASCADE Technology (OCCT), not the geometry kernel itself
\citep{cadquery}.
FreeCAD also uses OCCT, but provides a CAD application with native model
history, geometry-constrained sketches, and a Python API \citep{freecad}.
BenchCAD's CadQuery tasks emphasize parametric program synthesis, whereas
our tasks require agents to construct and save native FreeCAD documents
with the specified feature structure. Both are CAD workflows; our setting
additionally exercises the application-level document and feature system.

\paragraph{CAD-bench: functional CAD generation.}
CAD-bench, presented at an ICML 2026 workshop, executes generated CAD code or
agent-produced STEP files and checks the exported geometry with task-specific
tests, including dimensional and pose verification, thread-profile analysis,
and rigid-body simulation in Blender \citep{cadbenchfunctional}. Its 17 tasks
span four difficulty tiers, from basic solids to threaded mating pairs and
gear trains. It therefore tests whether generated parts function together,
whereas \bench\ requires native parametric FreeCAD documents across 100 tasks
and scores their geometric similarity and compliance with named dimensions
without simulating mechanisms. Despite the similar name, it is a separate
benchmark with a different evaluation target.

\paragraph{3DCodeBench: mesh-based procedural 3D authoring.}
3DCodeBench evaluates Blender Python scripts generated from text and images,
including iterative execution and visual feedback \citep{threedcodebench}.
Its evaluated outputs are meshes, assessed through executability,
geometric and perceptual similarity, and pairwise human preferences.
In contrast, \bench\ targets engineering-oriented mechanical CAD:
native parametric FreeCAD documents with boundary-representation (B-rep)
geometry and task-specific dimensional and structural checks.
B-rep geometry represents surfaces and their topological relationships,
rather than only a polygonal surface approximation \citep{occtmodelingdata}.
Thus the distinction is between procedural mesh assets and dimensionally
specified, editable CAD models, not between imprecise and precise software.
3DCodeBench includes manufactured objects, but its evaluation does not
require native mechanical-CAD feature histories or engineering-dimension
compliance. Likewise, our checks do not certify complete manufacturability.

\paragraph{Terminal-Bench and Harbor: agent evaluation infrastructure.}
Terminal-Bench provides complementary methodology for evaluating agents
on difficult terminal workflows \citep{terminalbench}.
Our benchmark is built on Harbor, a framework for specifying and evaluating
agent tasks in sandboxed environments \citep{harborframework}.
We also contributed three FreeCAD tasks to Terminal-Bench 3.0/4.0:
two create-and-edit tasks (\code{freecad-impeller} and
\code{freecad-spring-clip}) and one image-to-CAD task
(\code{freecad-platform-drawing}) \citep{terminalbenchfreecadtasks}.
These tasks follow the same create-and-edit and engineering-drawing
reconstruction workflow types evaluated in \bench.

\section{Tasks and Dataset Construction}
\subsection{Task workflows}
Table~\ref{tab:workflows} describes the released task composition.

\begin{table}[ht]
\centering
\small
\begin{tabularx}{\linewidth}{@{}l r X X@{}}
\toprule
Workflow & Tasks & Agent input & Required CAD artifacts\\
\midrule
Text-to-CAD Creation & 30 & Part description and specified parameters
  & Native parametric model\\
Text-to-CAD Create-and-Edit & 30 & Base description, edit instruction, and target requirements
  & Base and edited native models\\
Image-to-CAD & 40 & Engineering drawing and general execution requirements
  & Native model reconstructed from the drawing\\
\bottomrule
\end{tabularx}
\caption{The v3 task composition. Text-to-CAD create-and-edit is a two-stage construction
workflow, not editing an externally supplied CAD document.}
\label{tab:workflows}
\end{table}

In create-and-edit, the agent first constructs the base from text and then
produces the edited model; both artifacts are graded. The task does not provide
an externally authored starting \fcstd. Scoring the two artifacts does not
establish that the agent actually reused the base or preserved its feature
history. Image-to-CAD tasks provide engineering drawings rather than the
held-back dimensional specification.

\subsection{Example tasks}
\begin{figure}[htbp]
\centering
\includegraphics[width=\linewidth,trim=170.4bp 1.7bp 170.4bp 26.6bp,clip]{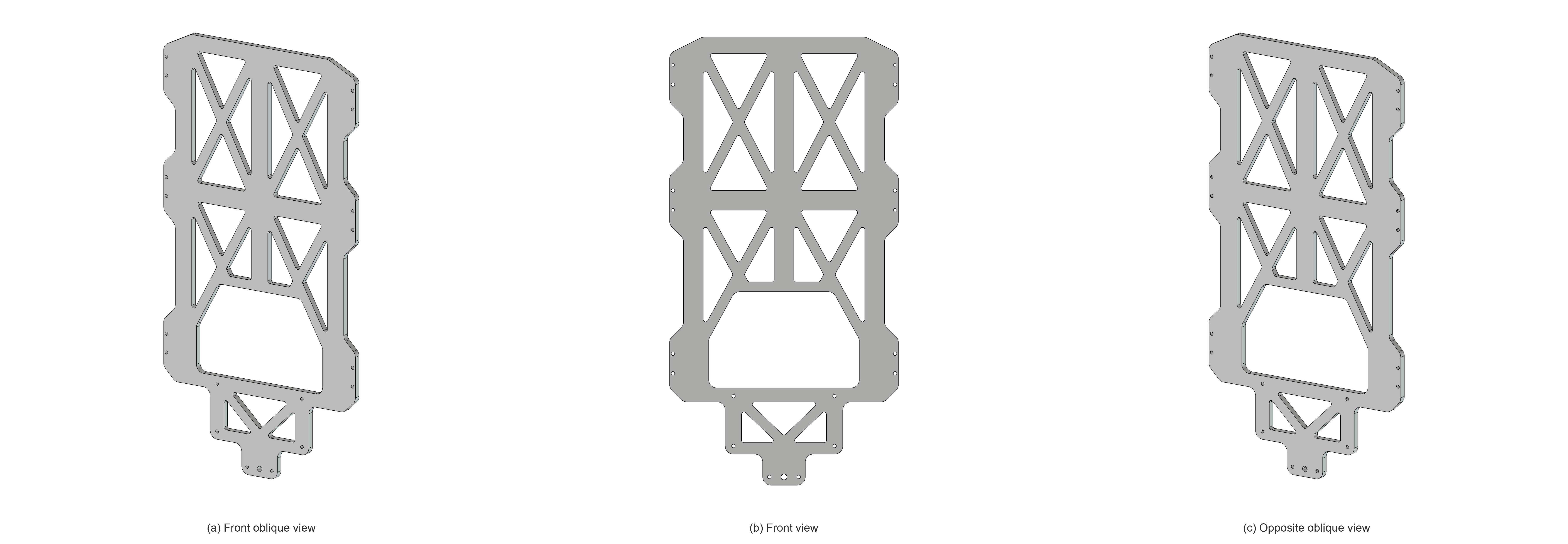}
\caption{Grader reference of text-to-CAD creation example \code{freecad-2002df3ce7}, a
trussed upper plate. (b) faces the plate, showing its 18 internal openings,
side mounting holes, and centered end extension.}
\label{fig:example-plate}
\end{figure}

\paragraph{Text-to-CAD creation examples.}
Two creation tasks illustrate the range of specifications.
Figure~\ref{fig:example-plate} shows \code{freecad-2002df3ce7}, a trussed
upper plate: a thin planar frame whose outline is the union of three
rectangular sections with rounded corner chamfers, with mirrored trapezoidal
recesses on the sides of the largest section. Diagonal truss webs separate
seventeen triangular openings and one central six-sided opening, forming 18
internal loops; peripheral mounting holes and a center bore complete the part.
Its 86 key parameters fall into five groups (Table~\ref{tab:example-groups}).
The description fixes the conventions needed to place them, including a sketch
frame rotated about global $+X$ so that layout $+y$ points along global $-Z$,
and centroid, base-length, and base-angle definitions for each isosceles
opening. The grader reference builds the plate from one sketch and one pad; it
is one solid with 203 faces, 280\,mm wide, 549\,mm tall, and 8\,mm thick.

\begin{figure}[htbp]
\centering
\includegraphics[width=\linewidth,trim=66bp 1.7bp 66bp 125.8bp,clip]{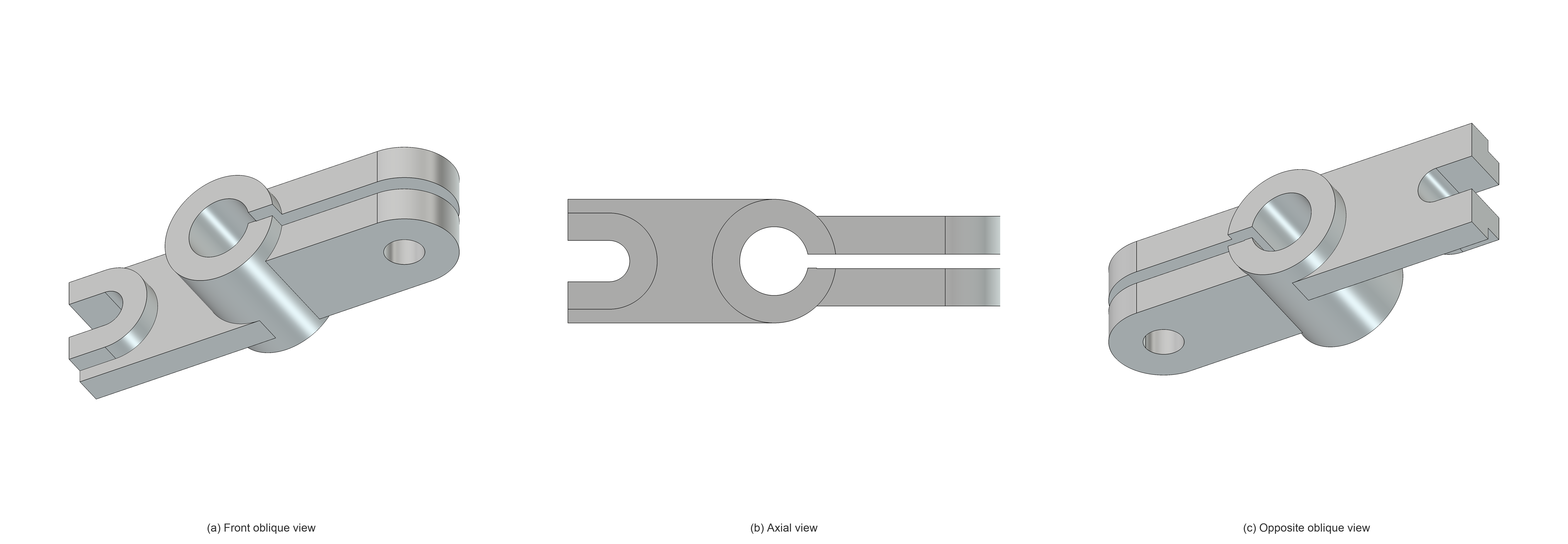}
\caption{Grader reference of text-to-CAD creation example \code{freecad-fa0a1abef3}, an
asymmetric clevis joint connector. (b) looks along the split-ring axis,
showing the fork, the large clevis, and the slotted side arm.}
\label{fig:example-clevis}
\end{figure}

Figure~\ref{fig:example-clevis} shows \code{freecad-fa0a1abef3}, an
asymmetric clevis joint connector: a rectangular central body with a large
split-ring clevis at its $+X$ end, a smaller fork at its $-X$ end, and a long
rounded side arm with three relief slots. The description states that the two
ends are not mirror images and fixes global axes along, across, and through
the connector. Its 44 key parameters fall into eight groups. The grader
reference uses eight sketches, four pads, and four pockets; it is one solid
with 31 faces, measuring 157\,mm by 45\,mm by 50\,mm.
Appendix~\ref{app:example-spec} reproduces both complete descriptions and
parameter lists as given to the agent.

The held-back grader specification also states how each parameter is checked.
Only 8 of the plate's 86 parameters and 7 of the connector's 44 are bound to
witnesses on the final geometry; the task's \code{param_check.py} measures the
rest from native sketch and feature geometry. Neither task uses task-specific
bindings. A candidate therefore earns specification credit only when its
model exposes the named dimensions in a form these checks can measure.
Section~\ref{sec:example-outcome} reports how the evaluated systems fared on
both tasks.

\begin{table}[htbp]
\centering
\small
\textit{(a) Trussed upper plate, \code{freecad-2002df3ce7}}\\[0.3em]
\begin{tabular}{@{}lrrrr@{}}
\toprule
Parameter group & Parameters & Witness & Custom & Native\\
\midrule
Plate outline, thickness, and side recesses & 21 & 2 & 0 & 19\\
Truss layout and diagonal angles & 7 & 1 & 0 & 6\\
Mounting holes and center bore & 15 & 2 & 0 & 13\\
Triangular and central openings & 41 & 3 & 0 & 38\\
Second right-triangular opening & 2 & 0 & 0 & 2\\
\midrule
Total & 86 & 8 & 0 & 78\\
\bottomrule
\end{tabular}
\\[0.8em]
\textit{(b) Asymmetric clevis joint connector, \code{freecad-fa0a1abef3}}\\[0.3em]
\begin{tabular}{@{}lrrrr@{}}
\toprule
Parameter group & Parameters & Witness & Custom & Native\\
\midrule
Central body & 3 & 1 & 0 & 2\\
Large split-ring clevis & 10 & 4 & 0 & 6\\
Clevis pin-hole cut & 2 & 0 & 0 & 2\\
Fork prongs & 5 & 1 & 0 & 4\\
Fork U-shaped opening & 5 & 0 & 0 & 5\\
Side arm & 9 & 0 & 0 & 9\\
Central relief slot & 3 & 1 & 0 & 2\\
Upper and lower relief slots & 7 & 0 & 0 & 7\\
\midrule
Total & 44 & 7 & 0 & 37\\
\bottomrule
\end{tabular}

\caption{Key-parameter groups of the two text-to-CAD creation examples and how the grader
checks them. Witness: bound to a witness on the final geometry. Custom: a
task-specific binding applied by \code{param_check.py}. Native: measured by
\code{param_check.py} from native sketch or feature geometry.}
\label{tab:example-groups}
\end{table}

\begin{figure}[htbp]
\centering
\begin{tabular}{@{}cc@{}}
\includegraphics[width=0.27\linewidth,trim=69.1bp 65bp 43bp 41.8bp,clip]{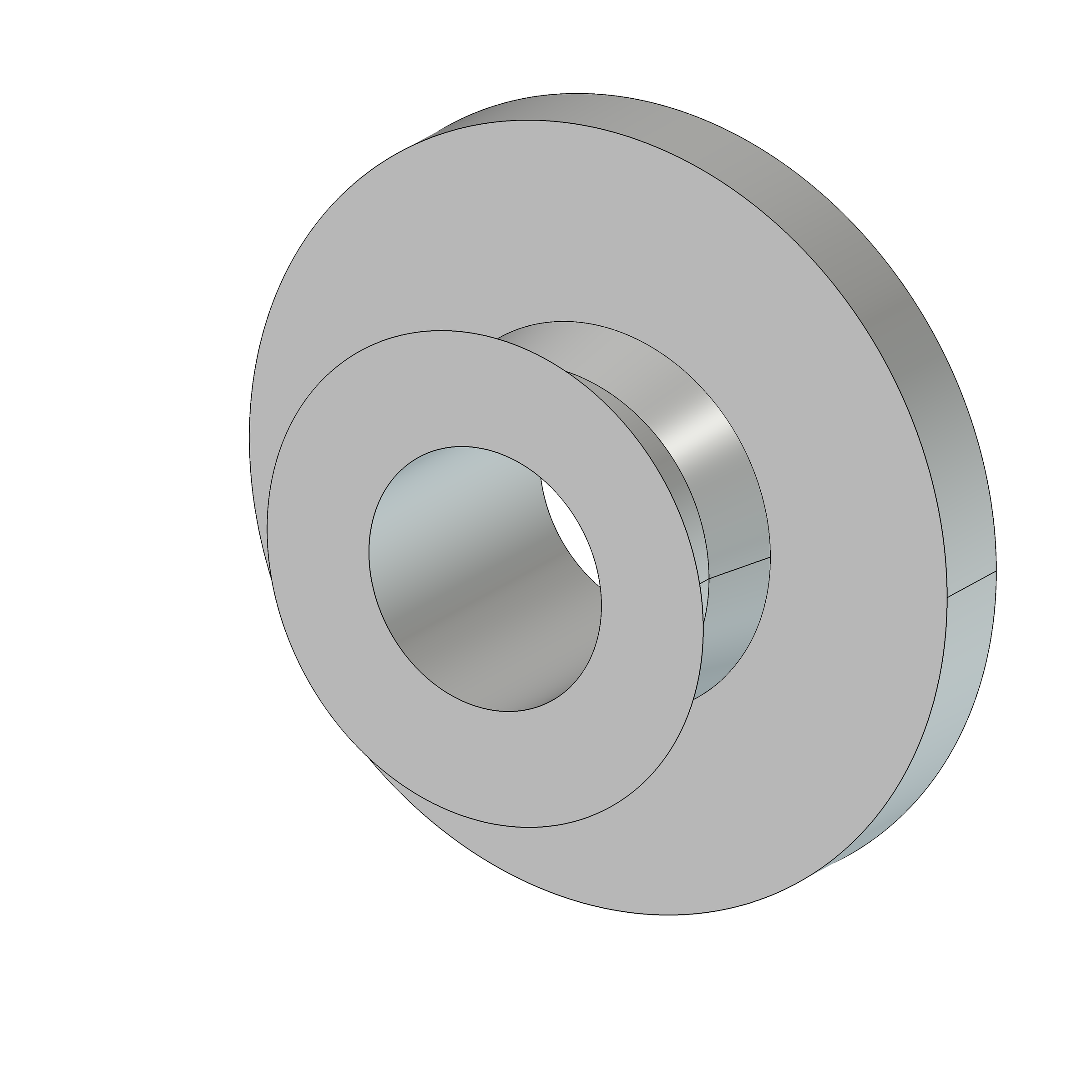} &
\includegraphics[width=0.27\linewidth,trim=61bp 61bp 61bp 61bp,clip]{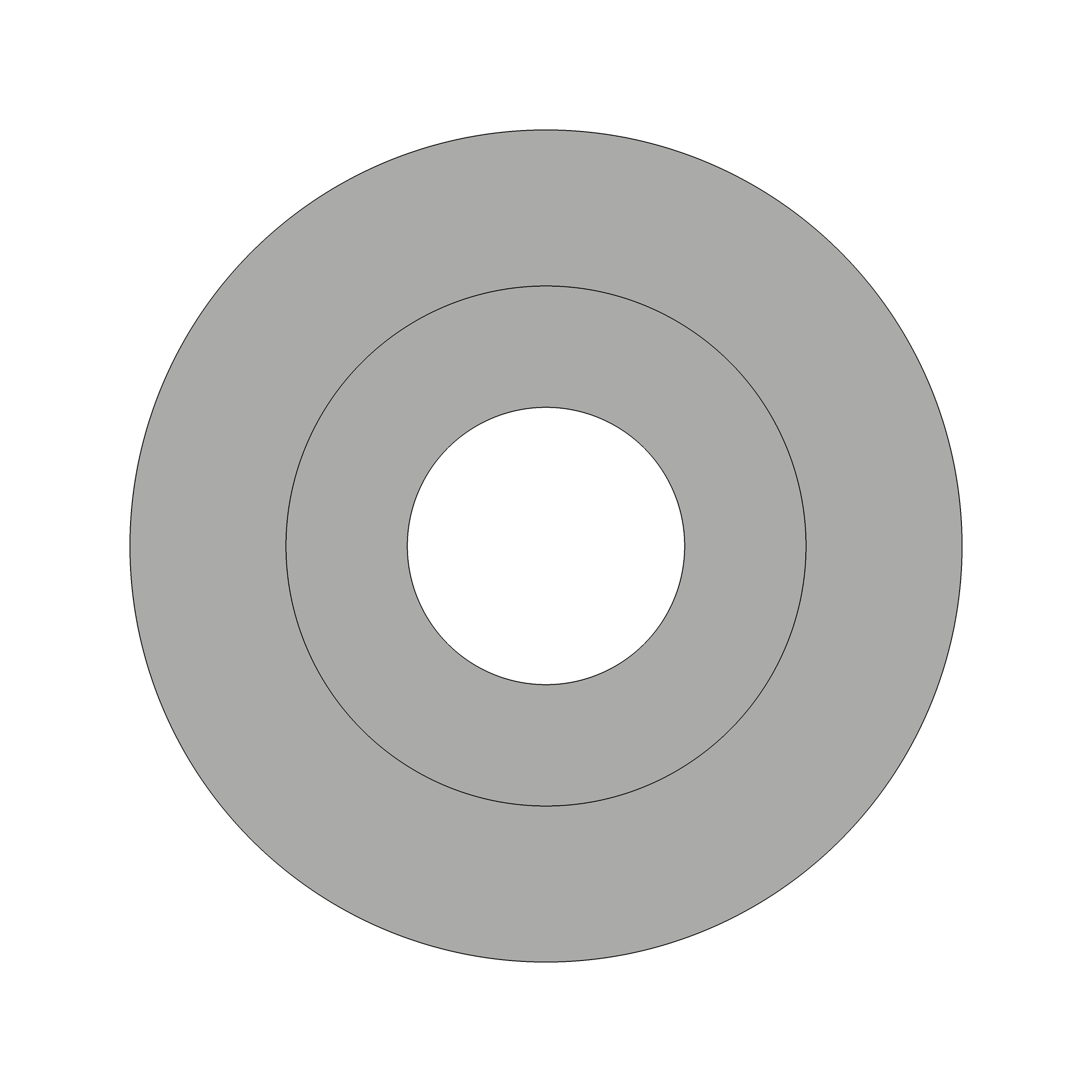}\\
\small (a) Base: oblique view & \small (b) Base: axial view\\[0.8em]
\includegraphics[width=0.27\linewidth,trim=69.1bp 65bp 43bp 41.8bp,clip]{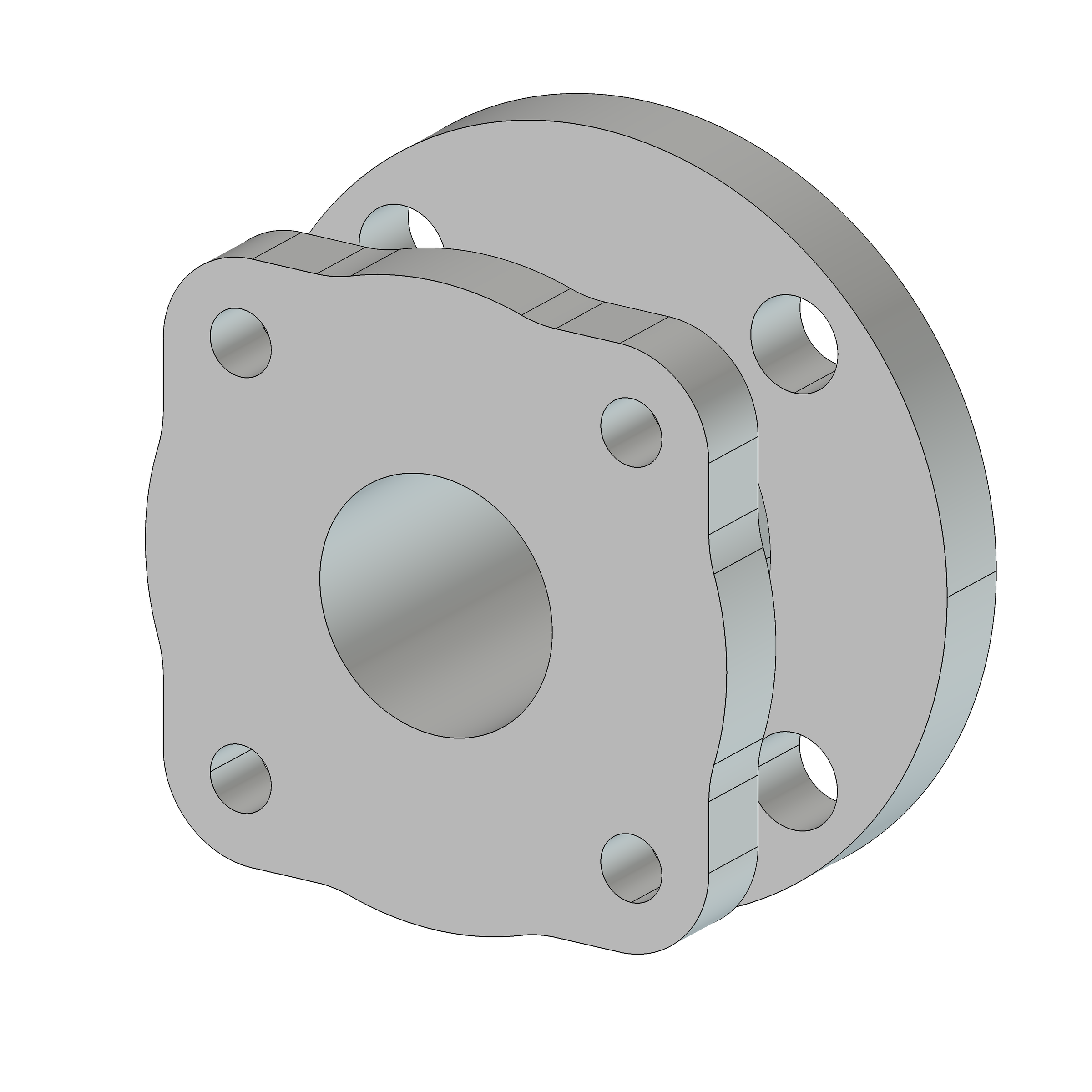} &
\includegraphics[width=0.27\linewidth,trim=61bp 61bp 61bp 61bp,clip]{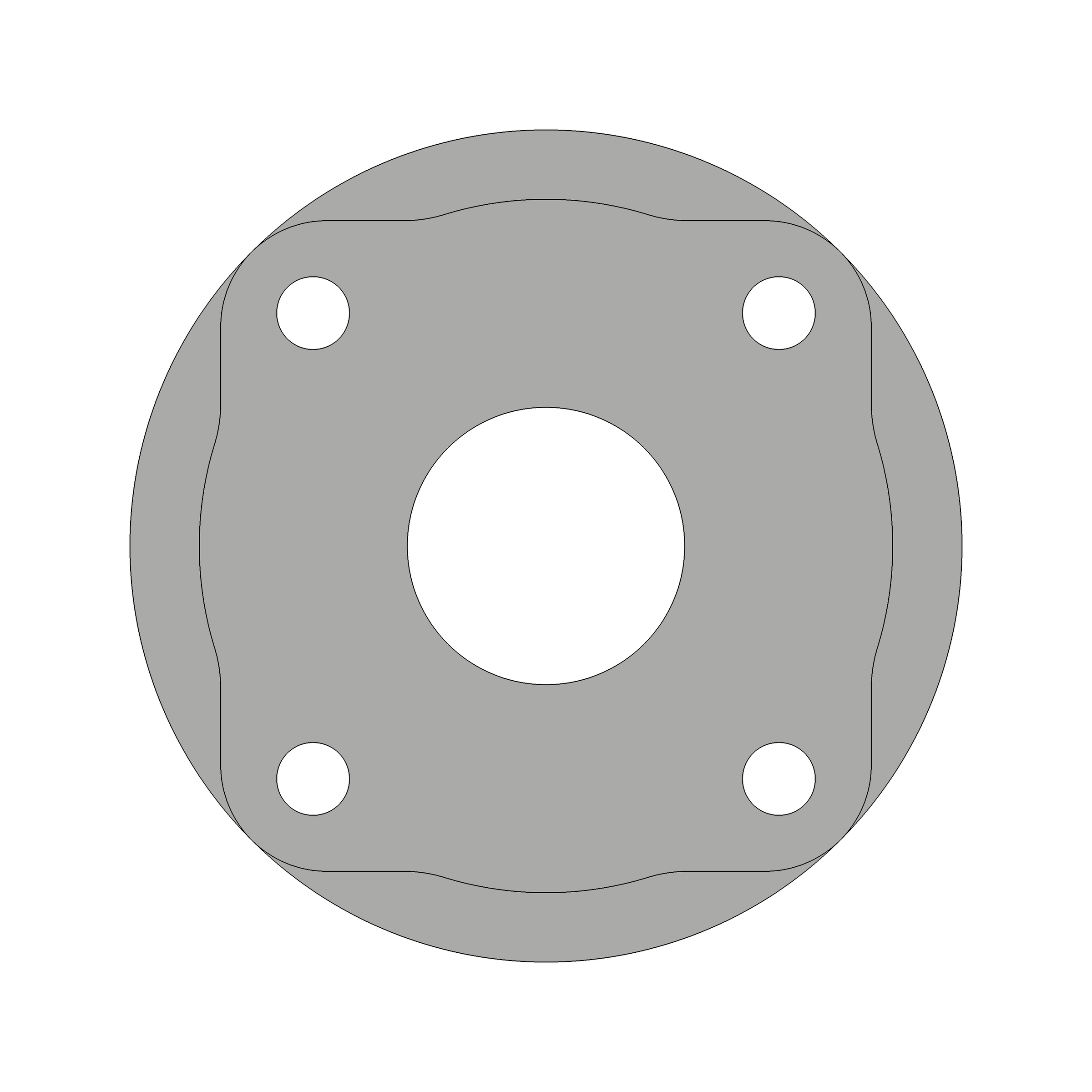}\\
\small (c) Edited target: oblique view & \small (d) Edited target: axial view
\end{tabular}
\caption{Text-to-CAD create-and-edit example \code{freecad-ff7883ab0c}: trusted base
reference (top) and edited target reference (bottom), shown in oblique and
axial views. Each view uses the same angle and scale for both stages. The edit
adds the front flange plate, its four bolt holes, and four rear-face openings
while retaining the axisymmetric body and central bore.}
\label{fig:example-edit-task}
\end{figure}

\paragraph{Text-to-CAD create-and-edit example.}
Figure~\ref{fig:example-edit-task} illustrates \code{freecad-ff7883ab0c}.
The agent first builds an axisymmetric flanged inlet housing: a continuous
50.8\,mm central bore, a tapered inlet rim, a tubular middle body, a tapered
rear hub, and a plain 152.4\,mm rear flange disk. The edit adds a
fourfold-symmetric front plate, 16.51\,mm thick, whose outline joins side,
corner, and transition arcs with short tangent connectors. It also adds four
13.34\,mm through bolt holes in the plate and four 19.05\,mm openings cut
34.54\,mm deep from the rear flange face, each pattern placed from a signed
seed center at uniform angular spacing. The tubular body, inlet rim, rear hub,
rear disk envelope, and central passage must remain unchanged. Fifteen base
parameters and 14 additional target parameters specify diameters, axial
positions, arc radii, hole sizes and counts, and seed-center offsets. The
complete supplied base description, edit instruction, and parameter lists
appear in Appendix~\ref{app:example-edit-spec}.

This task couples an axisymmetric base with a non-circular, arc-bounded plate
and two four-hole patterns on opposite faces. The script must save both
\code{answer_base.FCStd} and \code{answer_edit.FCStd}, each as an editable
single-body PartDesign model. Both stages are evaluated; a plausible final
silhouette alone does not demonstrate correct base construction or compliance
with the plate-outline and hole requirements. As with the other
create-and-edit tasks, the two saved artifacts do not by themselves prove that
the target reused the base feature history.

\begin{figure}[htbp]
\centering
\includegraphics[width=0.8\linewidth,trim=268bp 71bp 250bp 24bp,clip]{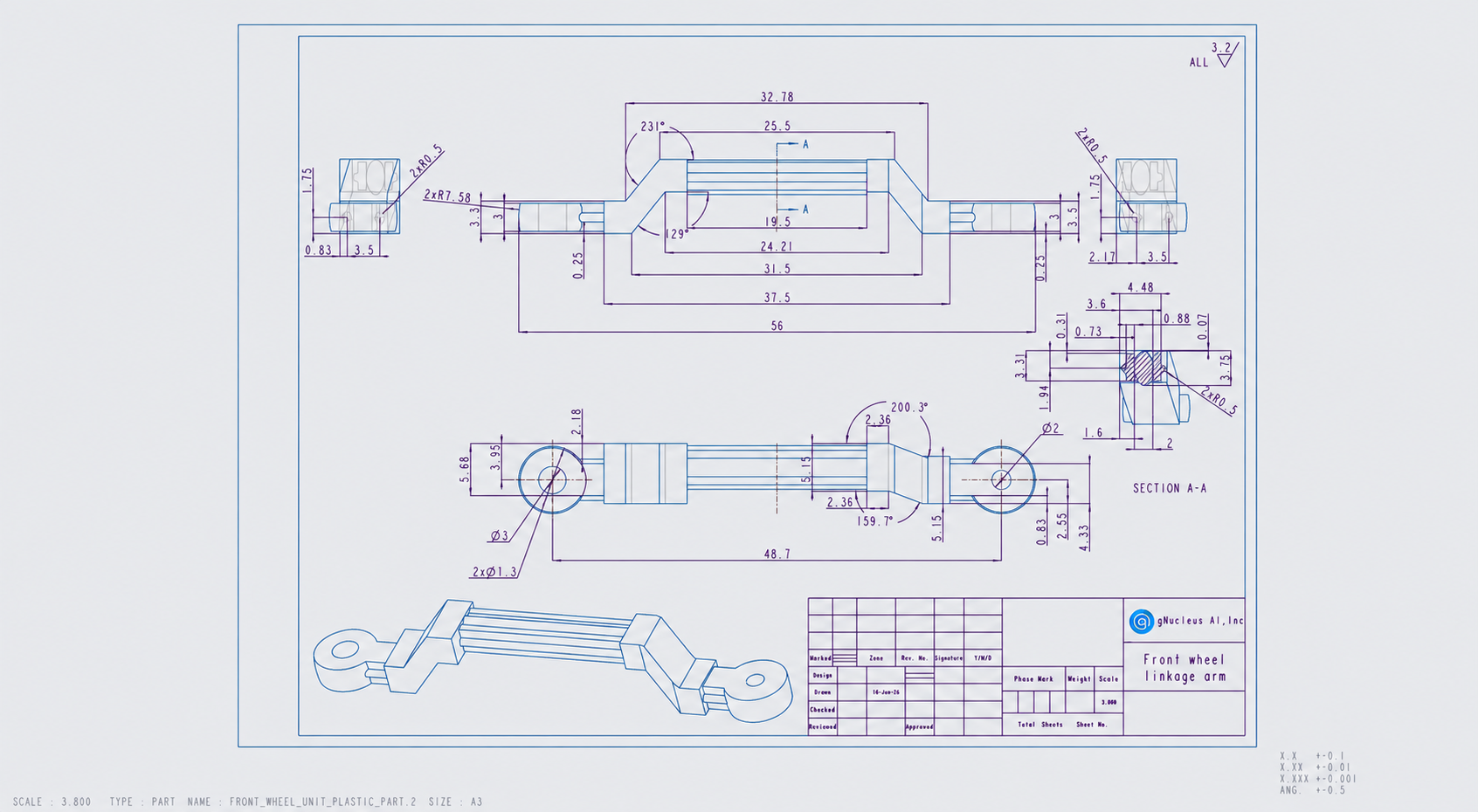}\\[0.8em]
\includegraphics[width=0.6\linewidth,trim=31.7bp 192bp 42.5bp 210.2bp,clip]{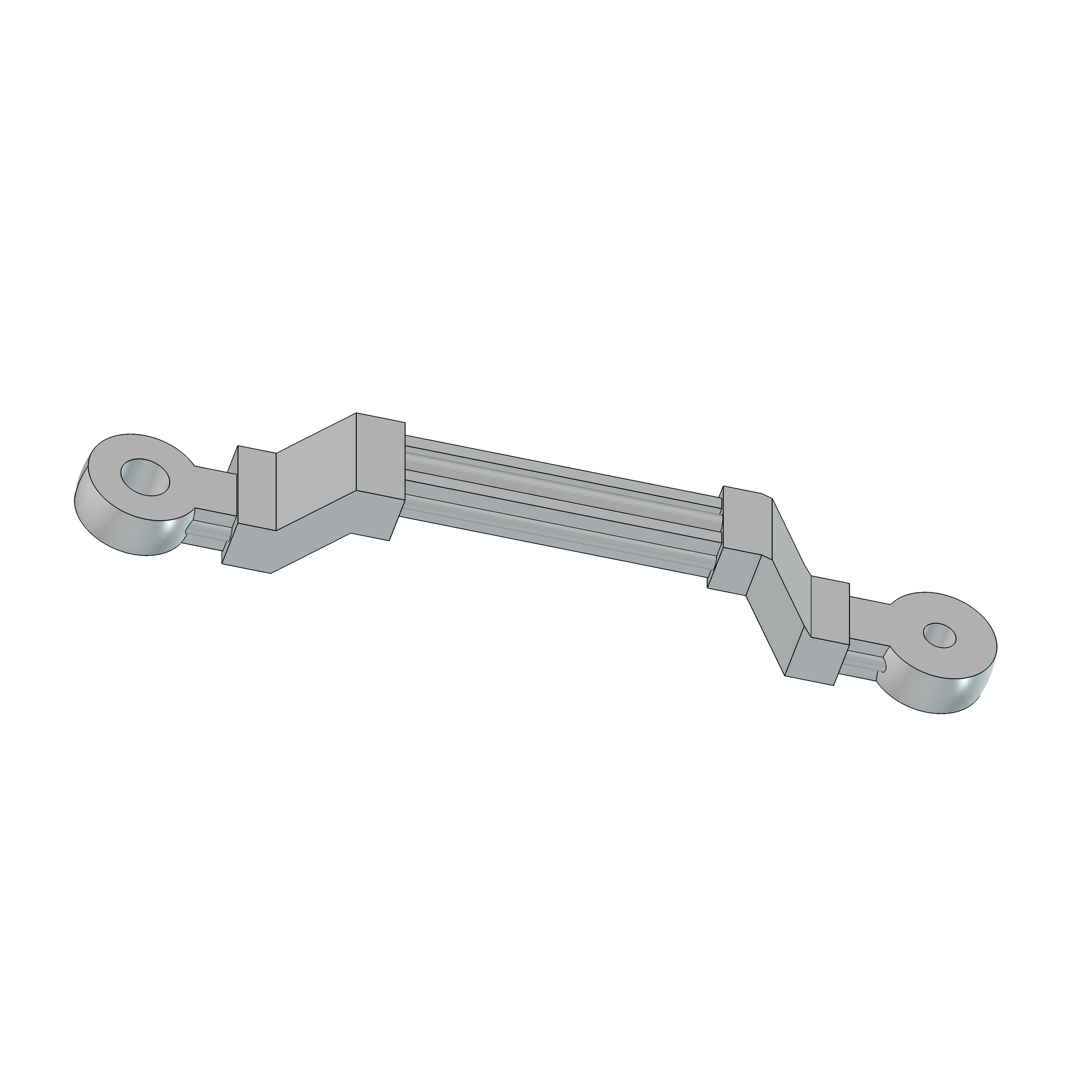}
\caption{Image-to-CAD example \code{freecad-9749149002}, a front-wheel
linkage arm. Top: the engineering drawing supplied to the agent, cropped to
its frame; the scale line and general-tolerance note outside the frame are
not shown. Bottom: an oblique view of the held-back grader reference.}
\label{fig:example-image-task}
\end{figure}

\paragraph{Image-to-CAD example.}
Figure~\ref{fig:example-image-task} shows the input drawing and trusted
reference for \code{freecad-9749149002}. The drawing combines orthographic,
section, and isometric views of a linkage arm with two eye ends, angled
transitions, and ribbed connecting beams. Dimension labels specify its
56\,mm overall length, 19.5\,mm depth, unequal 3\,mm and 2\,mm eye holes,
and local radii and offsets. The agent must reconcile these views and
dimensions to construct an editable single-solid PartDesign model.

Unlike the text-based examples, the agent receives no separate prose
description or key-parameter list: the drawing is its only geometry input.
It must write \code{/app/answer.py} and save \code{/app/answer.FCStd}
under FreeCAD 1.1.0. The reference shown here is held back during evaluation;
it illustrates the intended output, not an additional agent input.

\subsection{Reference-model statistics}
We count solid-modeling steps using FreeCAD's native feature type hierarchy:
objects derived from \code{Part::Feature}, excluding body containers,
sketches, and datum helpers. Intermediate and hidden feature-history objects
are included; a pattern counts as one feature regardless of instance count.
Generic \code{Part::Feature} objects count too, so this statistic alone
does not certify parametric editability. Sketch counts are reported separately.
The main distribution gives each task one observation, using the target
reference for create-and-edit rather than counting both stages as separate tasks.
The 100 final reference models contain 687 solid-modeling feature objects in total, with a mean of 6.87, a median of 6, and a range of 1--20 features per task. For the 30 create-and-edit pairs, the base and target means are 2.47 and 5.60, respectively. These describe the provided reference constructions; they do not establish a minimum number of features required for a valid alternative solution or a calibrated measure of task difficulty.
\par
\begin{table}[ht]
\centering\small
\begin{tabular}{@{}lrrrrr@{}}
\toprule
Workflow & Tasks & Mean features & Median & Min & Max\\
\midrule
All final references & 100 & 6.87 & 6 & 1 & 20\\
Text-to-CAD Creation & 30 & 5.20 & 4 & 1 & 20\\
Text-to-CAD Create-and-Edit (target) & 30 & 5.60 & 5.5 & 2 & 10\\
Image-to-CAD & 40 & 9.08 & 8 & 2 & 20\\
\bottomrule
\end{tabular}
\caption{Measured solid-modeling feature counts in the v3 grader references. Text-to-CAD create-and-edit contributes its target only to the 100-task distribution.}
\label{tab:feature-statistics}
\end{table}

\begin{figure}[htbp]
\centering
\includegraphics[width=\linewidth]{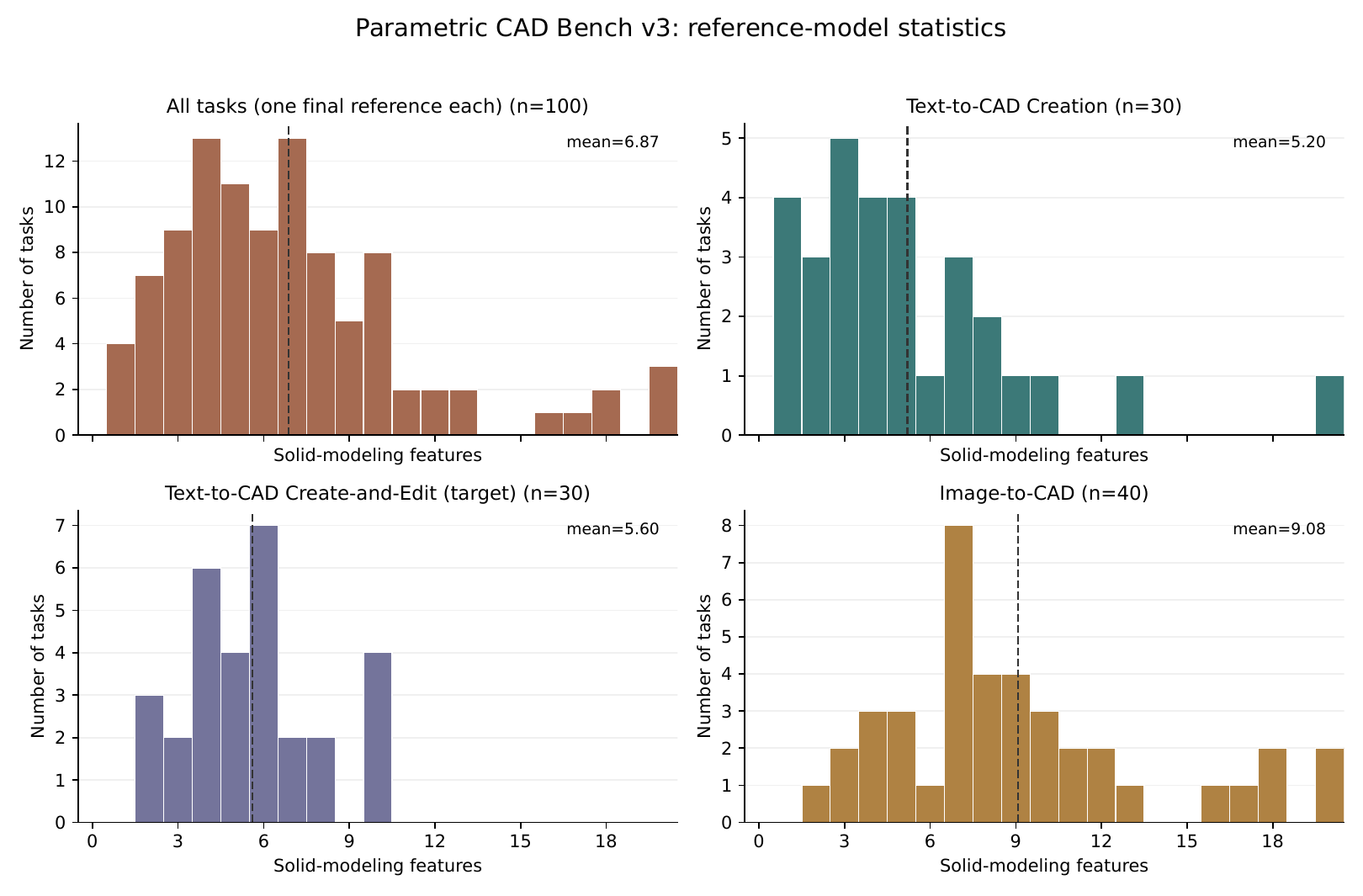}
\caption{Solid-modeling features per final reference model. The aggregate
contains 100 tasks, not 130 documents; dashed lines mark cohort means.
Counts reflect reference construction choices rather than a calibrated
difficulty measure or a required solution sequence.}
\label{fig:feature-counts}
\end{figure}

Figure~\ref{fig:feature-counts} shows unit-bin histograms for the full benchmark
and each workflow. Supplementary figures show feature-type composition, sketch
counts, and base/target distributions for the 30 edit pairs.

\section{Execution and Evaluation}
\subsection{Pinned runtime and isolation}
The runtime uses FreeCAD 1.1.0, Harbor 0.23.0, and task schema 1.3.
The verifier pins \code{gnucleus-freecad-validator} 0.6.0 with its v2 scorer.
The benchmark version, validator package version, and scorer version are
distinct identifiers. Reference geometry and task-specific grader assets are
placed in a separate verifier environment rather than the agent environment.
Each public job records its agent version and concurrency.

\subsection{Programmatic scoring}
The verifier computes two independent axes, \code{geometry_similarity}
and \code{cad_spec_consistency}, using the default v2 scorer in
\code{gnucleus-freecad-validator} 0.6.0 \citep{freecadvalidator060}.

\paragraph{Geometric similarity.}
The scorer compares the candidate and reference solids using surface-type
areas, volume, total surface area, normalized principal moments of inertia,
and face-center iterative closest point (ICP) alignment.
Let $q_{\mathrm{surf}},q_{\mathrm{vol}},q_{\mathrm{area}},q_{\mathrm{mom}}$
and $q_{\mathrm{ICP}}$ denote their respective subscores in $[0,1]$.
After the gates below pass, the geometric score is
\begin{align}
p &= \frac{0.05q_{\mathrm{surf}}+0.175q_{\mathrm{vol}}
     +0.175q_{\mathrm{area}}+0.10q_{\mathrm{mom}}}{0.50},\\
g &= p\left(0.50+0.50q_{\mathrm{ICP}}\right).
\label{eq:geometry}
\end{align}
Volume receives full credit at relative error at most $0.1\%$ and zero
subscore at $1\%$ or more; the corresponding thresholds for surface-type
areas, total surface area, and normalized principal moments are $1\%$ and
$10\%$. Intermediate errors receive logarithmically interpolated credit.
Surface-type comparison uses the largest per-type relative area error,
with a denominator floor of $1\%$ of the larger total surface area.
An independently measured OCCT oriented bounding box is a gate, not an
additional weighted subscore: maximum relative error across sorted extents
of $10\%$ or more sets $g=0$.
Structural gates also require one non-empty \code{PartDesign::Body}
producing one solid through allowed native features rather than baked
shape holders. Candidates with more than 5,000 faces, or face/vertex counts
differing from the reference by more than $50\%$ of the larger count, are
rejected. Gate failures override the spatial multiplier's nonzero floor.
Face-center alignment depends on face decomposition, so congruent geometry
built through a different feature history need not receive $g=1$.

\paragraph{CAD specification consistency.}
The scorer parses the task's held-back specification and checks its
parameters against measurements extracted from the candidate document.
CAD engineers wrote the parameter-checking logic shipped in each task's
\code{tests/grader/param_check.py}. Most tasks share a workflow-level
implementation, and a subset adds case-specific checks. These checks encode
ways to measure the dimensions, the correct placement of
dimensions on the CAD part, feature relationships, and derived quantities
required by the task's specification. They supplement generic measurements
with engineering-specific checks beyond global shape similarity.
The v2 path honors trusted geometry bindings and case-local
\code{param_check.py} refinements where supplied; remaining checks use
generic per-kind measurements. The default scalar relative tolerance is
$1\%$, with case-specific measurement logic and binding tolerances where
applicable. Let $n$ be the number of parsed specification parameters and
$f$ the number classified as inconsistent or not found. The default v2
failure budget is 10:
\begin{equation}
s =
\begin{cases}
0, & n=0,\\[2pt]
\max\!\left(0,1-\dfrac{f}{\min(n,10)}\right), & n>0.
\end{cases}
\label{eq:specification}
\end{equation}
For $n\geq10$, each failed or unmeasurable parameter costs 0.1, preventing
large specifications from diluting failures. This score measures the
encoded checks, not every possible engineering requirement or functional
editability under arbitrary future changes.

\paragraph{Combined task reward.}

Let $g,s\in[0,1]$ denote geometry and specification scores. Define
\begin{equation}
H(g,s)=
\begin{cases}
0, & g=0\ \text{or}\ s=0,\\[2pt]
\dfrac{2gs}{g+s}, & \text{otherwise}.
\end{cases}
\label{eq:harmonic}
\end{equation}
For creation and image-to-CAD, the task reward is $r=H(g,s)$.
For create-and-edit, each output is scored separately, then combined:
\begin{equation}
r_{\mathrm{edit}} =
H\!\left(H(g_{\mathrm{base}},s_{\mathrm{base}}),
         H(g_{\mathrm{target}},s_{\mathrm{target}})\right).
\label{eq:edit}
\end{equation}
Thus a zero score on a required axis or deliverable cannot be rescued by
another axis.
The wrappers clamp rewards to $[0,1]$; text reward files use six decimal
places, and the image-to-CAD wrapper also rounds its JSON reward to six
decimal places.

The reported overall percentage is
\begin{equation}
S = \frac{100}{N}\sum_{i=1}^{N}r_i,\qquad N=100.
\label{eq:overall}
\end{equation}
Final exception outcomes contribute zero. Workflow percentages are computed
within each cohort, so $S=0.30S_{\mathrm{create}}+
0.30S_{\mathrm{edit}}+0.40S_{\mathrm{image}}$.

This continuous reward is not Pass@1. We additionally report the fraction of
final rewards exactly equal to 1.0 as the operational perfect-task rate.
A scored trial is one without a final exception; scored count and
positive-reward count are different quantities.

\section{Results and Analysis}
\subsection{Dated leaderboard snapshot}
Figures~\ref{fig:overall}--\ref{fig:image} visualize the 13-row September 24,
2026 snapshot. The plots are generated from the published leaderboard rows.
The overall chart shows 95\% intervals computed as the mean reward $\pm$
$1.96\,s/\sqrt{100}$, where $s$ is the sample standard deviation of a
system's 100 per-task rewards. They describe uncertainty over the task sample,
given one final trial per task, not variability across repeated runs of a
system. Workflow charts show point estimates only.

Claude Code with Claude Opus 5.5 achieves the highest overall point estimate,
61.03\%, followed by mini-swe-agent with the same model at 57.96\%, Codex with
GPT-6 Astra at 56.87\%, and Claude Code with Claude Fable 5.1 at 56.75\%.
Point-estimate ordering alone does not demonstrate a statistically decisive
ranking.

The workflow leaders differ: Fable 5.1 with Claude Code has the highest creation
point estimate (55.48\%), Astra with Codex leads create-and-edit (44.37\%), and
Opus 5.5 with Claude Code leads image-to-CAD (84.66\%). Because workflows use
different task populations, these differences cannot be attributed solely to
input modality.

\subsection{Same-model and same-agent comparisons}
We pair final rewards by task ID, rather than comparing only cohort means.
For the following descriptive counts, two outcomes are \emph{similar} when
their rewards differ by at most 0.05 (5 percentage points); a higher outcome
must exceed this tolerance. Similarity is not a statistical equivalence claim,
nor does it imply success: two zero-reward outcomes also count as similar.
Final exception outcomes retain their zero rewards.

\paragraph{Opus 5.5 versus Opus 5 with Claude Code.}
Within Claude Code, the overall point estimate rises from
50.86\% for Opus 5 to 61.03\% for Opus 5.5, a 10.18 percentage-point increase.
The release analysis reports a paired 95\% confidence interval of
5.22--15.13 points for this improvement \citep{parametriccadbenchv3opus55}.
It is a normal approximation over the 100 per-task reward differences: their
mean $\pm$ $1.96\,s_d/\sqrt{100}$, where $s_d$ is their sample standard
deviation.
Table~\ref{tab:opus-model-paired} shows that this is not a uniform improvement
across workflows. For creation, Opus 5.5 is higher on 8 tasks, similar on 18,
and lower on 4; for create-and-edit, the corresponding counts are 4, 23, and 3.
Image-to-CAD shows the broadest improvement: Opus 5.5 is higher on 22 of 40
tasks, similar on 15, and lower on only 3. Its median paired drawing-task gain
is 7.62 points, and all five drawings receiving zero reward under Opus 5
receive positive reward under Opus 5.5. The 22.02-point image-to-CAD mean gain
accounts for approximately 87\% of the overall improvement after applying
the benchmark's 30/30/40 workflow weights.

\paragraph{Claude Code versus mini-swe-agent with Opus 5.5.}
To support multimodal tasks, we add an explicit system prompt to the
mini-swe-agent run: when a task references an image, the agent must first
send that image through its multimodal observation channel, inspect the
resulting visual input, and only then proceed with the task. This is paired
with a configured bridge that converts marked, base64-encoded shell output
into image content for the model; the prompt alone does not enable image
transport. The Opus 5.5 configuration uses mini-swe-agent 2.4.6, preserves
complete image payloads, and leaves malformed image output as text.

With Opus 5.5 held constant, Claude Code exceeds mini-swe-agent by 3.08 points.
However, the paired interval computed the same way is $[-1.15,7.31]$ points, so
these results do not establish an overall advantage for either agent.
The workflow-level comparison in Table~\ref{tab:opus-agent-paired} is more
informative than the overall ordering. Claude Code leads creation by
8.07 points, with 9 higher outcomes, 20 similar outcomes, and 1 lower outcome.
mini-swe-agent leads create-and-edit by 1.68 points: Claude Code is higher on
2 tasks, similar on 24, and lower on 4. For image-to-CAD, 30 of 40 outcomes
are similar, and each agent is higher on 5 tasks. Claude Code's 2.90-point
mean advantage there therefore reflects the magnitudes of the differing
outcomes, rather than a greater number of task-level wins. Both systems
exceed 80\% mean reward on drawing reconstruction with direct image input.
These comparisons remain conditional on the evaluated configurations and
dates, including different Claude Code versions; they are not completely
controlled model or agent ablations.

\begin{table}[htbp]
\centering\small
\begin{tabular}{lrrrrrr}
\toprule
Workflow & $N$ & A mean & B mean & A higher & Similar & B higher \\
\midrule
Text-to-CAD Creation & 30 & 50.87\% & 49.48\% & 8 & 18 & 4 \\
Text-to-CAD Create-and-Edit & 30 & 39.69\% & 36.52\% & 4 & 23 & 3 \\
Image-to-CAD & 40 & 84.66\% & 62.64\% & 22 & 15 & 3 \\
Overall & 100 & 61.03\% & 50.86\% & 34 & 56 & 10 \\
\bottomrule
\end{tabular}
\caption{Claude Code held constant; A: Opus 5.5, B: Opus 5. Higher means a task reward advantage greater than 5 percentage points; Similar includes the boundary. These counts do not establish statistical equivalence.}
\label{tab:opus-model-paired}
\end{table}
\begin{table}[htbp]
\centering\small
\begin{tabular}{lrrrrrr}
\toprule
Workflow & $N$ & A mean & B mean & A higher & Similar & B higher \\
\midrule
Text-to-CAD Creation & 30 & 50.87\% & 42.79\% & 9 & 20 & 1 \\
Text-to-CAD Create-and-Edit & 30 & 39.69\% & 41.38\% & 2 & 24 & 4 \\
Image-to-CAD & 40 & 84.66\% & 81.77\% & 5 & 30 & 5 \\
Overall & 100 & 61.03\% & 57.96\% & 16 & 74 & 10 \\
\bottomrule
\end{tabular}
\caption{Opus 5.5 held constant; A: Claude Code, B: mini-swe-agent with multimodal bridge. Higher means a task reward advantage greater than 5 percentage points; Similar includes the boundary. These counts do not establish statistical equivalence.}
\label{tab:opus-agent-paired}
\end{table}

\paragraph{Initial takeaway.}
The clearest observed gain is in engineering-drawing reconstruction:
Opus 5.5 improves Claude Code's image-to-CAD score by 22.02 points, versus
1.39 points for creation and 3.17 points for create-and-edit.
Both Opus 5.5 systems exceed 80\% on image-to-CAD (84.66\% and 81.77\%),
so the strong drawing results are not confined to one agent implementation
\citep{parametriccadbenchv3opus55}.
This suggests substantial progress in drawing-to-CAD task completion,
but not uniformly reliable CAD authoring: the same systems reach only
39.69\% and 41.38\% on create-and-edit. These are distinct task cohorts,
not a controlled test of modality alone.
Together with the launch post's Gemini comparison across two agents,
the results also indicate that model capability and agent interface both
matter; they do not identify a universally superior agent
\citep{parametriccadbenchv3}.

\subsection{Multimodal CAD performance}
Image-to-CAD produces the largest separation between the evaluated
model--agent systems. Across the 13 rows, its mean reward ranges from
8.18\% to 84.66\%, a 76.49-point spread. By comparison, creation ranges
from 23.32\% to 55.48\% (32.16 points), and create-and-edit from 23.23\%
to 44.37\% (21.14 points). The five highest image-to-CAD systems achieve
62.64--84.66\%, whereas the remaining eight achieve only 8.18--31.10\%.
Drawing reconstruction therefore distinguishes these systems much more
sharply than either text-based workflow.

This pattern is also visible within a single agent family. Across the three
Claude Code rows, image-to-CAD ranges from 62.64\% with Opus 5 to 84.66\%
with Opus 5.5, a 22.02-point difference. The corresponding spreads are
only 6.00 points for creation and 3.17 points for create-and-edit.
Agent configuration matters as well: Gemini 3.8 Flash achieves 31.10\%
on image-to-CAD with mini-swe-agent's multimodal bridge, versus 17.45\%
with Antigravity, a 13.65-point difference. Its creation and create-and-edit
differences between these agents are smaller, at 4.66 and 5.90 points,
respectively.

These results suggest that multimodal CAD task completion is a major
differentiator among the evaluated systems. Strong drawing performance
requires interpreting views, dimensions, and geometric relationships and
translating them into a valid parametric model. However, these scores measure
that entire workflow, not visual perception in isolation: CAD construction,
tool use, image transport, and failure recovery can all affect the outcome.
The larger score spread does not by itself establish which of these factors
causes the gap, particularly because the workflows contain different tasks.

\subsection{Efficiency}
\label{sec:efficiency}
\paragraph{Cost per score point.}
We divide the known API cost of the complete 100-task run by its overall
score on the 0--100 scale. API costs come from the usage recorded in each
job's artifacts. For Grok Build with Grok 4.7 and for mini-swe-agent with
Kimi K3, Muse Spark 1.3, and Claude Opus 5.5, whose artifacts record no or
incomplete cost, they come from our audits, which price the token usage in
the saved trajectories at standard API rates. This descriptive ratio measures
expenditure per achieved score point, not the marginal cost of improving a
system by one point. Codex with GPT-6 Astra has the lowest ratio among all 13
systems:
\$317.81 divided by 56.87 points, or \$5.59 per point. Its total spend is
not the lowest, but the additional expenditure buys substantially higher
reward: compared with GPT-5.6 Sol using Codex, Astra costs only \$25.82
more while scoring 26.89 points higher.

\paragraph{Per-task speed.}
Astra also has the shortest mean elapsed trial duration, 14.42 minutes,
compared with 16.58 minutes for Sol, 31.11 for Claude Code with Fable 5.1,
and 32.51 for Claude Code with Opus 5.5. These are final-trial elapsed times,
including setup, agent execution, and verification, rather than isolated
model inference latency. We do not use whole-job wall time to rank speed,
because concurrency, pauses, retries, and scheduling differ between runs.

\paragraph{Reward--cost trade-off.}
Astra's 56.87\% mean reward is close to Fable 5.1's 56.75\%, but its known
API cost is approximately 70\% lower and its mean trial duration is less
than half as long. The higher-scoring Claude Code--Opus 5.5 system improves
the point estimate by 4.17 points over Astra, at 1.81 times the API cost
and 2.25 times the mean trial duration. Astra therefore offers a favorable
observed balance of reward, cost, and per-task speed, rather than the highest
reward at any price. These comparisons describe the evaluated configurations
and their audited API accounting; they exclude some compute and hosting costs
and do not establish why one system requires less time or expenditure.

\subsection{Observed task difficulty}
Across the 1,300 selected final trials from the 13 evaluated systems,
393 (30.23\%) earn zero reward. Table~\ref{tab:observed-difficulty} summarizes
the pattern by workflow. Create-and-edit has the largest zero-outcome rate,
41.54\%, compared with 22.56\% for creation and 27.50\% for image-to-CAD.
These counts use the official final outcomes, including zero-valued final
exceptions and reruns of provider or platform failures that produced no model
outcome, rather than the superseded attempts.

\begin{table}[htbp]
\centering\small
\begin{tabular}{lrrrr}
\toprule
Workflow & Zero outcomes & All 13 zero & $\geq$7 zero & Any perfect \\
\midrule
Text-to-CAD Creation & 22.56\% & 0 & 5 & 6 \\
Text-to-CAD Create-and-Edit & 41.54\% & 5 & 11 & 3 \\
Image-to-CAD & 27.50\% & 0 & 8 & 6 \\
Overall & 30.23\% & 5 & 24 & 15 \\
\bottomrule
\end{tabular}
\caption{Observed difficulty across 13 evaluated systems. Zero outcomes is the fraction of final trials with zero reward. The last three columns count distinct tasks: zero in all 13 systems, zero in at least 7 systems, and reward exactly 1.0 in at least one system. Cohorts contain 30 text-to-CAD creation, 30 text-to-CAD create-and-edit, and 40 image-to-CAD tasks.}
\label{tab:observed-difficulty}
\end{table}

Difficulty is concentrated in a subset of tasks. Twenty-four tasks receive
zero reward from at least 7 of the 13 systems: 5 creation tasks, 11
create-and-edit tasks, and 8 image-to-CAD tasks. Five tasks receive zero
from every system, all in create-and-edit. On 26 tasks, no system reaches
0.5: 9 creation, 16 create-and-edit, and 1 image-to-CAD. At the other end,
15 tasks earn a perfect reward of exactly 1.0 from at least one system
(6 creation, 3 create-and-edit, and 6 image-to-CAD). Thus, strong aggregate
drawing performance coexists with substantial remaining difficulty,
especially in two-stage CAD construction.
These are descriptive outcomes for the evaluated systems and configurations,
not estimates of a task's inherent solvability.

The saved diagnostics also show that zero reward is usually associated
with the CAD scoring criteria rather than a final execution exception.
Of the 393 zeros, 326 involve a zero geometry component, including 50
that also have a zero specification component. Another 33 fail only
specification consistency; the remainder comprise 23 final exceptions,
7 missing outputs, and 4 scorer crashes. This breakdown describes observed
failure surfaces, not independently adjudicated root causes.

\paragraph{Example task outcomes.}
\label{sec:example-outcome}
The two creation examples fail in different ways. The trussed plate
(Figure~\ref{fig:example-plate}) is one of the 24 tasks with zero reward from
at least 7 systems: 9 of the 13 systems receive zero
(Table~\ref{tab:example-trials-plate}). Eight of these zeros come from the
specification axis alone. These outcomes have geometry similarity between
0.126 and 0.445, but 12 to 80 of the 86 parameters are inconsistent or not
found, which exhausts the failure budget of 10 in
Equation~\ref{eq:specification}. The ninth zero is a final exception. Among
the four positive outcomes, Claude Fable 5.1 with Claude Code is nearly exact
(0.999, with every parameter consistent); the others miss 1, 7, and 9
parameters. Codex with GPT-5.6 Terra earns 0.140 with lower geometry
similarity (0.232) than five zero-reward attempts, because it stays within the
failure budget. On this task, specification consistency rather than overall
shape separates the rewarded attempts.

The clevis connector (Figure~\ref{fig:example-clevis}) shows the opposite
pattern (Table~\ref{tab:example-trials-clevis}). Twelve of the 13 systems earn
positive reward, and Claude Opus 5 with Claude Code earns 1.000 to three
decimals, with every parameter consistent. Only Codex with GPT-5.6 Terra
receives zero, on both axes.
The other eleven positive outcomes miss at most 7 of the 44 parameters, but
their geometry similarity ranges only from 0.175 to 0.500. Two of them, Claude
Fable 5.1 with Claude Code and Gemini 3.8 Flash with mini-swe-agent, pass every
parameter check yet earn 0.419 because their geometry similarity is 0.265.
Here, geometry rather than the named dimensions limits most rewards.

\begin{table}[htbp]
\centering
\small
\begin{tabular}{@{}llrrrrl@{}}
\toprule
Model & Agent & Reward & $g$ & $s$ & Failed & Outcome\\
\midrule
Claude Fable 5.1 & Claude Code & 0.999 & 0.998 & 1.000 & 0 & positive\\
Grok 4.6 & Grok Build & 0.628 & 0.482 & 0.900 & 1 & positive\\
GPT-6 Astra & Codex & 0.375 & 0.500 & 0.300 & 7 & positive\\
GPT-5.6 Terra & Codex & 0.140 & 0.232 & 0.100 & 9 & positive\\
Claude Opus 5 & Claude Code & 0.000 & 0.146 & 0.000 & 13 & specification zero\\
Claude Opus 5.5 & Claude Code & 0.000 & 0.445 & 0.000 & 12 & specification zero\\
Claude Opus 5.5 & mini-swe-agent & 0.000 & --- & --- & --- & final exception\\
GPT-5.6 Sol & Codex & 0.000 & 0.299 & 0.000 & 80 & specification zero\\
Gemini 3.8 Flash & Antigravity & 0.000 & 0.195 & 0.000 & 19 & specification zero\\
Gemini 3.8 Flash & mini-swe-agent & 0.000 & 0.247 & 0.000 & 17 & specification zero\\
Grok 4.7 & Grok Build & 0.000 & 0.319 & 0.000 & 18 & specification zero\\
Kimi K3 & mini-swe-agent & 0.000 & 0.286 & 0.000 & 19 & specification zero\\
Muse Spark 1.3 & mini-swe-agent & 0.000 & 0.126 & 0.000 & 24 & specification zero\\
\bottomrule
\end{tabular}

\caption{Final outcomes of the 13 evaluated systems on the trussed-plate
example \code{freecad-2002df3ce7}. $g$ and $s$ are the geometry and
specification scores; Failed counts the parameters, of 86, that are
inconsistent or not found. ---: no scored output after a final exception.}
\label{tab:example-trials-plate}
\end{table}

\begin{table}[htbp]
\centering
\small
\begin{tabular}{@{}llrrrrl@{}}
\toprule
Model & Agent & Reward & $g$ & $s$ & Failed & Outcome\\
\midrule
Claude Opus 5 & Claude Code & 1.000 & 1.000 & 1.000 & 0 & positive\\
Claude Opus 5.5 & Claude Code & 0.615 & 0.500 & 0.800 & 2 & positive\\
Claude Fable 5.1 & Claude Code & 0.419 & 0.265 & 1.000 & 0 & positive\\
Gemini 3.8 Flash & mini-swe-agent & 0.419 & 0.265 & 1.000 & 0 & positive\\
GPT-6 Astra & Codex & 0.397 & 0.264 & 0.800 & 2 & positive\\
Kimi K3 & mini-swe-agent & 0.397 & 0.264 & 0.800 & 2 & positive\\
Muse Spark 1.3 & mini-swe-agent & 0.397 & 0.264 & 0.800 & 2 & positive\\
Grok 4.7 & Grok Build & 0.307 & 0.190 & 0.800 & 2 & positive\\
Gemini 3.8 Flash & Antigravity & 0.299 & 0.190 & 0.700 & 3 & positive\\
Claude Opus 5.5 & mini-swe-agent & 0.252 & 0.218 & 0.300 & 7 & positive\\
Grok 4.6 & Grok Build & 0.252 & 0.218 & 0.300 & 7 & positive\\
GPT-5.6 Sol & Codex & 0.243 & 0.175 & 0.400 & 6 & positive\\
GPT-5.6 Terra & Codex & 0.000 & 0.000 & 0.000 & 26 & geometry and specification zero\\
\bottomrule
\end{tabular}

\caption{Final outcomes of the 13 evaluated systems on the clevis-connector
example \code{freecad-fa0a1abef3}. Columns as in
Table~\ref{tab:example-trials-plate}; Failed counts the parameters, of 44,
that are inconsistent or not found.}
\label{tab:example-trials-clevis}
\end{table}

\section{Discussion and Limitations}
The benchmark is a curated 100-task FreeCAD evaluation, not a representative
sample of all engineering CAD. Family dependencies and selection choices
limit generalization. Native single-body PartDesign requirements do not cover
general assembly mating, FEM/FEA, manufacturing qualification, or broad
organic-surface modeling.

Create-and-edit measures two-stage construction, not maintenance of an
externally authored document. Grader rewards are heuristic proxies:
geometric similarity and checked parameters do not certify all future edits,
design intent, or engineering validity. Alternative valid constructions may
challenge reference-based comparisons.

Public assets, providers' undisclosed training data, and changing provider implementations
limit contamination claims and reproducibility. Agent tools, prompts,
serving paths, and budgets confound model-only comparisons. Single official
attempts do not establish run-to-run robustness. Known API costs exclude some
compute and hosting costs.

\section{Conclusion}
\bench\ evaluates agentic CAD authoring across text-based creation,
create-and-edit, and engineering-drawing reconstruction. Its 100 tasks require
native parametric FreeCAD models, evaluated through geometric similarity and
CAD-engineer-curated specification checks.

Across 13 evaluated model--agent systems, the largest performance differences
occur in image-to-CAD. Opus 5.5's improvement over Opus 5 is concentrated in
drawing reconstruction, while agent comparisons reveal workflow-dependent
strengths rather than a universally superior interface. Create-and-edit remains
challenging: five tasks receive zero reward from every evaluated system.
Cost and speed also distinguish systems; Codex with GPT-6 Astra offers the
lowest observed API cost per score point and the shortest mean trial duration.

These results highlight substantial progress alongside remaining limitations
in reliable parametric CAD construction. By releasing versioned tasks,
execution configurations, scoring code, and inspectable run artifacts,
Parametric CAD Bench provides a versioned, inspectable foundation for measuring
progress in engineering-oriented CAD agents.

\bibliographystyle{plainnat}
\bibliography{references}

\appendix
\section{Reproducibility and Artifact Index}
The following public resources accompany this paper:
\begin{itemize}
  \item Benchmark: \url{https://cadbench.ai/}
  \item V3 tasks and leaderboard:
    \href{https://hub.harborframework.com/datasets/gnucleus-ai/cad-bench/v3?tab=leaderboard&leaderboard=v3}{\nolinkurl{https://hub.harborframework.com/datasets/gnucleus-ai/cad-bench/v3}}
  \item Run configurations and trial artifacts: the public Harbor job linked
    from each leaderboard row.
  \item Scoring code (validator): \url{https://github.com/gNucleus-AI/freecad-validator}
\end{itemize}
The frozen dataset digest is
\begin{quote}\small
\nolinkurl{sha256:8a786725d3648a76282fc5868df06da3d5e00d603f283573a53430803465d78a}.
\end{quote}
The reported runs use Harbor 0.23.0, FreeCAD 1.1.0, and
\code{gnucleus-freecad-validator} 0.6.0, published on PyPI and tagged
\code{v0.6.0} (commit \code{2a5d67f}) in the validator repository. The trials
of the linked Harbor jobs include agent logs, generated models, and saved
verifier outputs.

The benchmark tasks, including their specifications, reference models, and
drawings, were authored by gNucleus AI and are released under the Apache
License 2.0, as is the validator. The task images redistribute FreeCAD and
Open CASCADE Technology under the LGPL-2.1, with their license notices.

Apart from the four audited API costs described in
Section~\ref{sec:efficiency}, every reported statistic is computed from the
public tasks and the final trials of the public Harbor jobs, and can be
recomputed from them.

\section{Text-to-CAD Creation Example Specifications}
\label{app:example-spec}
For each creation example, the agent receives the description and key
parameters below, reproduced exactly as given, including their full decimal
precision. The instruction adds the output contract shared by creation tasks:
write a FreeCAD 1.1.0 script to \code{/app/answer.py} that saves
\code{/app/answer.FCStd}, building the model as a parametric feature tree in a
single PartDesign Body that produces exactly one solid, with editable features
rather than a pre-computed shape.

\subsection{Trussed upper plate (\code{freecad-2002df3ce7})}
The key parameters are listed in Table~\ref{tab:example-parameters-2002df3ce7}.

\begin{quote}
\small
A trussed upper plate consists of a thin, planar outer frame with rounded corners, repeated diagonal truss webs, open cutouts, peripheral mounting holes, and a centered end extension.

The outer profile is centered on \texttt{truss\_\allowbreak{}layout\_\allowbreak{}symmetry\_\allowbreak{}axis\_\allowbreak{}x} and is the union of three rectangular sections: large\_outer\_rectangle, middle\_outer\_rectangle, and small\_outer\_rectangle. The large rectangle has two mirrored trapezoidal side recesses on each side, with equal diagonal flanks, and a centered lower flat bounded by symmetric rounded chamfers. The two upper corners of the large rectangle have chamfer setback \texttt{large\_\allowbreak{}outer\_\allowbreak{}rectangle\_\allowbreak{}upper\_\allowbreak{}corner\_\allowbreak{}chamfer\_\allowbreak{}setback}. All rectangular-section corner chamfers use \texttt{rectangular\_\allowbreak{}section\_\allowbreak{}corner\_\allowbreak{}chamfer\_\allowbreak{}fillet\_\allowbreak{}radius}.

The truss layout is centered on \texttt{truss\_\allowbreak{}layout\_\allowbreak{}symmetry\_\allowbreak{}axis\_\allowbreak{}x}. All named layout x and y coordinates use a sketch frame at the global origin. Rotate this frame from the global XY plane by -\texttt{base\_\allowbreak{}sketch\_\allowbreak{}plane\_\allowbreak{}rotation\_\allowbreak{}angle} about global +X, using the right-hand convention. Thus layout +x points along global +X, layout +y points along global -Z, and the plate extends from global Y=0 in the +Y direction by \texttt{plate\_\allowbreak{}thickness}. Its triangular substructures are six large triangles, six medium triangles, two right triangles, one small triangle, and two chamfer-cut right triangles. Each strictly isosceles triangle is defined by its centroid, ideal base length, and base angle; its rounded profile uses \texttt{truss\_\allowbreak{}cutout\_\allowbreak{}transition\_\allowbreak{}radius}. large\_triangular\_opening\_pair\_01 through \_03 and medium\_triangular\_opening\_pair\_01 through \_02 each define two mirrored isosceles openings. The first and third large pairs have exterior vertical bases, while the second has interior vertical bases. The first medium pair has lower horizontal bases and the second has upper horizontal bases. medium\_triangular\_opening\_pair\_03 has an interior vertical base and two diagonal flanks meeting at a rounded exterior apex. An upper horizontal truncation joins the interior base to the upper flank; all four transitions are rounded with \texttt{truss\_\allowbreak{}cutout\_\allowbreak{}transition\_\allowbreak{}radius}. The two right triangles and the small triangle have lower horizontal bases and upward apices. The chamfer-cut right triangles have upper horizontal bases, exterior vertical legs, and rounded diagonal cut corners.

central\_trapezoidal\_opening defines the remaining non-triangular internal loop; together with the seventeen triangular openings, it forms eighteen internal loops. It shares \texttt{truss\_\allowbreak{}layout\_\allowbreak{}symmetry\_\allowbreak{}axis\_\allowbreak{}x} with the outer profile and center bore. Its rounded six-sided boundary has an upper horizontal edge, two upper vertical sidewalls, two lower flanks sloping inward toward the bottom, and a lower horizontal edge. The lower flanks are inclined by \texttt{truss\_\allowbreak{}diagonal\_\allowbreak{}angle\_\allowbreak{}family\_\allowbreak{}3} from the vertical. The finished straight lower edge has \texttt{central\_\allowbreak{}trapezoidal\_\allowbreak{}opening\_\allowbreak{}lower\_\allowbreak{}flat\_\allowbreak{}width}. All six corners are tangent rounds of \texttt{central\_\allowbreak{}trapezoidal\_\allowbreak{}opening\_\allowbreak{}corner\_\allowbreak{}radius}; the named envelope and center locate the complete boundary. The plate body is governed by \texttt{plate\_\allowbreak{}thickness}. The triangular transitions use \texttt{truss\_\allowbreak{}cutout\_\allowbreak{}transition\_\allowbreak{}radius}, while central\_trapezoidal\_opening uses \texttt{central\_\allowbreak{}trapezoidal\_\allowbreak{}opening\_\allowbreak{}corner\_\allowbreak{}radius}. The mounting holes and center bore are governed by their respective diameters, pair spacings, rows, and axes.

\end{quote}

{\small
\setlength{\tabcolsep}{4pt}
\setlength{\LTcapwidth}{\linewidth}
\begin{longtable}{@{}l r l c@{}}
\caption{Key parameters of \texttt{freecad-2002df3ce7}, as given to the agent. Check: W, a witness on the final geometry; C, a custom binding; N, a native sketch or feature measurement by the task's \texttt{param\_check.py}.}\label{tab:example-parameters-2002df3ce7}\\
\toprule Parameter & Value & Unit & Check\\ \midrule \endfirsthead
\toprule Parameter & Value & Unit & Check\\ \midrule \endhead
\bottomrule \endlastfoot
\multicolumn{4}{@{}l}{\textit{Plate outline, thickness, and side recesses}}\\
\texttt{plate\_\allowbreak{}thickness} & 8.00 & mm & W\\
\texttt{base\_\allowbreak{}sketch\_\allowbreak{}plane\_\allowbreak{}rotation\_\allowbreak{}angle} & 90.0 & deg & N\\
\texttt{large\_\allowbreak{}outer\_\allowbreak{}rectangle\_\allowbreak{}width} & 280.00 & mm & N\\
\texttt{large\_\allowbreak{}outer\_\allowbreak{}rectangle\_\allowbreak{}height} & 447.00 & mm & N\\
\texttt{large\_\allowbreak{}outer\_\allowbreak{}rectangle\_\allowbreak{}center\_\allowbreak{}y} & $-$51.00 & mm & N\\
\texttt{middle\_\allowbreak{}outer\_\allowbreak{}rectangle\_\allowbreak{}width} & 144.00 & mm & N\\
\texttt{middle\_\allowbreak{}outer\_\allowbreak{}rectangle\_\allowbreak{}height} & 64.00 & mm & N\\
\texttt{middle\_\allowbreak{}outer\_\allowbreak{}rectangle\_\allowbreak{}center\_\allowbreak{}y} & 204.50 & mm & N\\
\texttt{small\_\allowbreak{}outer\_\allowbreak{}rectangle\_\allowbreak{}width} & 52.00 & mm & N\\
\texttt{small\_\allowbreak{}outer\_\allowbreak{}rectangle\_\allowbreak{}height} & 38.00 & mm & N\\
\texttt{small\_\allowbreak{}outer\_\allowbreak{}rectangle\_\allowbreak{}center\_\allowbreak{}y} & 255.50 & mm & N\\
\texttt{rectangular\_\allowbreak{}section\_\allowbreak{}corner\_\allowbreak{}chamfer\_\allowbreak{}fillet\_\allowbreak{}radius} & 10.00 & mm & W\\
\texttt{large\_\allowbreak{}outer\_\allowbreak{}rectangle\_\allowbreak{}lower\_\allowbreak{}corner\_\allowbreak{}chamfer\_\allowbreak{}flat\_\allowbreak{}width} & 200.00 & mm & N\\
\texttt{large\_\allowbreak{}outer\_\allowbreak{}rectangle\_\allowbreak{}upper\_\allowbreak{}corner\_\allowbreak{}chamfer\_\allowbreak{}setback} & 17.00 & mm & N\\
\texttt{large\_\allowbreak{}outer\_\allowbreak{}rectangle\_\allowbreak{}side\_\allowbreak{}trapezoidal\_\allowbreak{}recess\_\allowbreak{}depth} & 17.00 & mm & N\\
\texttt{large\_\allowbreak{}outer\_\allowbreak{}rectangle\_\allowbreak{}side\_\allowbreak{}trapezoidal\_\allowbreak{}recess\_\allowbreak{}01\_\allowbreak{}center\_\allowbreak{}y} & $-$149.50 & mm & N\\
\texttt{large\_\allowbreak{}outer\_\allowbreak{}rectangle\_\allowbreak{}side\_\allowbreak{}trapezoidal\_\allowbreak{}recess\_\allowbreak{}01\_\allowbreak{}outer\_\allowbreak{}height} & 98.28 & mm & N\\
\texttt{large\_\allowbreak{}outer\_\allowbreak{}rectangle\_\allowbreak{}side\_\allowbreak{}trapezoidal\_\allowbreak{}recess\_\allowbreak{}01\_\allowbreak{}inner\_\allowbreak{}flat\_\allowbreak{}height} & 47.72 & mm & N\\
\texttt{large\_\allowbreak{}outer\_\allowbreak{}rectangle\_\allowbreak{}side\_\allowbreak{}trapezoidal\_\allowbreak{}recess\_\allowbreak{}02\_\allowbreak{}center\_\allowbreak{}y} & 25.50 & mm & N\\
\texttt{large\_\allowbreak{}outer\_\allowbreak{}rectangle\_\allowbreak{}side\_\allowbreak{}trapezoidal\_\allowbreak{}recess\_\allowbreak{}02\_\allowbreak{}outer\_\allowbreak{}height} & 148.28 & mm & N\\
\texttt{large\_\allowbreak{}outer\_\allowbreak{}rectangle\_\allowbreak{}side\_\allowbreak{}trapezoidal\_\allowbreak{}recess\_\allowbreak{}02\_\allowbreak{}inner\_\allowbreak{}flat\_\allowbreak{}height} & 97.72 & mm & N\\
\multicolumn{4}{@{}l}{\textit{Truss layout and diagonal angles}}\\
\texttt{truss\_\allowbreak{}layout\_\allowbreak{}symmetry\_\allowbreak{}axis\_\allowbreak{}x} & 0.00 & mm & N\\
\texttt{truss\_\allowbreak{}cutout\_\allowbreak{}transition\_\allowbreak{}radius} & 3.00 & mm & W\\
\texttt{truss\_\allowbreak{}diagonal\_\allowbreak{}angle\_\allowbreak{}family\_\allowbreak{}1} & 27.3777028 & deg & N\\
\texttt{truss\_\allowbreak{}diagonal\_\allowbreak{}angle\_\allowbreak{}family\_\allowbreak{}2} & 63.4349488 & deg & N\\
\texttt{truss\_\allowbreak{}diagonal\_\allowbreak{}angle\_\allowbreak{}family\_\allowbreak{}3} & 28.8386275 & deg & N\\
\texttt{truss\_\allowbreak{}diagonal\_\allowbreak{}angle\_\allowbreak{}family\_\allowbreak{}4} & 46.1233027 & deg & N\\
\texttt{truss\_\allowbreak{}diagonal\_\allowbreak{}angle\_\allowbreak{}family\_\allowbreak{}5} & 165.8183429 & deg & N\\
\multicolumn{4}{@{}l}{\textit{Mounting holes and center bore}}\\
\texttt{mounting\_\allowbreak{}hole\_\allowbreak{}diameter} & 4.30 & mm & W\\
\texttt{center\_\allowbreak{}bore\_\allowbreak{}diameter} & 7.00 & mm & W\\
\texttt{center\_\allowbreak{}bore\_\allowbreak{}axis\_\allowbreak{}y} & 264.50 & mm & N\\
\texttt{center\_\allowbreak{}mounting\_\allowbreak{}hole\_\allowbreak{}pair\_\allowbreak{}half\_\allowbreak{}spacing} & 18.00 & mm & N\\
\texttt{upper\_\allowbreak{}inner\_\allowbreak{}mounting\_\allowbreak{}hole\_\allowbreak{}pair\_\allowbreak{}half\_\allowbreak{}spacing} & 62.00 & mm & N\\
\texttt{lower\_\allowbreak{}inner\_\allowbreak{}mounting\_\allowbreak{}hole\_\allowbreak{}pair\_\allowbreak{}half\_\allowbreak{}spacing} & 61.76 & mm & N\\
\texttt{upper\_\allowbreak{}inner\_\allowbreak{}mounting\_\allowbreak{}hole\_\allowbreak{}pair\_\allowbreak{}y} & 226.50 & mm & N\\
\texttt{lower\_\allowbreak{}inner\_\allowbreak{}mounting\_\allowbreak{}hole\_\allowbreak{}pair\_\allowbreak{}y} & 165.50 & mm & N\\
\texttt{side\_\allowbreak{}mounting\_\allowbreak{}hole\_\allowbreak{}axis\_\allowbreak{}x} & 136.00 & mm & N\\
\texttt{side\_\allowbreak{}mounting\_\allowbreak{}hole\_\allowbreak{}row\_\allowbreak{}1\_\allowbreak{}y} & 137.50 & mm & N\\
\texttt{side\_\allowbreak{}mounting\_\allowbreak{}hole\_\allowbreak{}row\_\allowbreak{}2\_\allowbreak{}y} & 113.50 & mm & N\\
\texttt{side\_\allowbreak{}mounting\_\allowbreak{}hole\_\allowbreak{}row\_\allowbreak{}3\_\allowbreak{}y} & $-$62.50 & mm & N\\
\texttt{side\_\allowbreak{}mounting\_\allowbreak{}hole\_\allowbreak{}row\_\allowbreak{}4\_\allowbreak{}y} & $-$86.50 & mm & N\\
\texttt{side\_\allowbreak{}mounting\_\allowbreak{}hole\_\allowbreak{}row\_\allowbreak{}5\_\allowbreak{}y} & $-$216.32 & mm & N\\
\texttt{side\_\allowbreak{}mounting\_\allowbreak{}hole\_\allowbreak{}row\_\allowbreak{}6\_\allowbreak{}y} & $-$240.32 & mm & N\\
\multicolumn{4}{@{}l}{\textit{Triangular and central openings}}\\
\texttt{large\_\allowbreak{}triangular\_\allowbreak{}opening\_\allowbreak{}pair\_\allowbreak{}01\_\allowbreak{}02\_\allowbreak{}base\_\allowbreak{}length} & 141.90 & mm & N\\
\texttt{large\_\allowbreak{}triangular\_\allowbreak{}opening\_\allowbreak{}pair\_\allowbreak{}01\_\allowbreak{}02\_\allowbreak{}base\_\allowbreak{}angle} & 27.3777028 & deg & N\\
\texttt{large\_\allowbreak{}triangular\_\allowbreak{}opening\_\allowbreak{}pair\_\allowbreak{}01\_\allowbreak{}positive\_\allowbreak{}member\_\allowbreak{}centroid\_\allowbreak{}x} & 86.75 & mm & N\\
\texttt{large\_\allowbreak{}triangular\_\allowbreak{}opening\_\allowbreak{}pair\_\allowbreak{}01\_\allowbreak{}positive\_\allowbreak{}member\_\allowbreak{}centroid\_\allowbreak{}y} & $-$170.50 & mm & N\\
\texttt{large\_\allowbreak{}triangular\_\allowbreak{}opening\_\allowbreak{}pair\_\allowbreak{}02\_\allowbreak{}positive\_\allowbreak{}member\_\allowbreak{}centroid\_\allowbreak{}x} & 24.25 & mm & N\\
\texttt{large\_\allowbreak{}triangular\_\allowbreak{}opening\_\allowbreak{}pair\_\allowbreak{}02\_\allowbreak{}positive\_\allowbreak{}member\_\allowbreak{}centroid\_\allowbreak{}y} & $-$170.50 & mm & N\\
\texttt{large\_\allowbreak{}triangular\_\allowbreak{}opening\_\allowbreak{}pair\_\allowbreak{}03\_\allowbreak{}positive\_\allowbreak{}member\_\allowbreak{}centroid\_\allowbreak{}x} & 86.78 & mm & N\\
\texttt{large\_\allowbreak{}triangular\_\allowbreak{}opening\_\allowbreak{}pair\_\allowbreak{}03\_\allowbreak{}positive\_\allowbreak{}member\_\allowbreak{}centroid\_\allowbreak{}y} & 16.50 & mm & N\\
\texttt{large\_\allowbreak{}triangular\_\allowbreak{}opening\_\allowbreak{}pair\_\allowbreak{}03\_\allowbreak{}base\_\allowbreak{}length} & 133.12 & mm & N\\
\texttt{large\_\allowbreak{}triangular\_\allowbreak{}opening\_\allowbreak{}pair\_\allowbreak{}03\_\allowbreak{}base\_\allowbreak{}angle} & 28.8386275 & deg & N\\
\texttt{medium\_\allowbreak{}triangular\_\allowbreak{}opening\_\allowbreak{}pair\_\allowbreak{}01\_\allowbreak{}02\_\allowbreak{}base\_\allowbreak{}length} & 73.49 & mm & N\\
\texttt{medium\_\allowbreak{}triangular\_\allowbreak{}opening\_\allowbreak{}pair\_\allowbreak{}01\_\allowbreak{}02\_\allowbreak{}base\_\allowbreak{}angle} & 62.6222972 & deg & N\\
\texttt{medium\_\allowbreak{}triangular\_\allowbreak{}opening\_\allowbreak{}pair\_\allowbreak{}01\_\allowbreak{}positive\_\allowbreak{}member\_\allowbreak{}centroid\_\allowbreak{}x} & 55.50 & mm & N\\
\texttt{medium\_\allowbreak{}triangular\_\allowbreak{}opening\_\allowbreak{}pair\_\allowbreak{}01\_\allowbreak{}positive\_\allowbreak{}member\_\allowbreak{}centroid\_\allowbreak{}y} & $-$230.85 & mm & N\\
\texttt{medium\_\allowbreak{}triangular\_\allowbreak{}opening\_\allowbreak{}pair\_\allowbreak{}02\_\allowbreak{}positive\_\allowbreak{}member\_\allowbreak{}centroid\_\allowbreak{}x} & 55.50 & mm & N\\
\texttt{medium\_\allowbreak{}triangular\_\allowbreak{}opening\_\allowbreak{}pair\_\allowbreak{}02\_\allowbreak{}positive\_\allowbreak{}member\_\allowbreak{}centroid\_\allowbreak{}y} & $-$110.15 & mm & N\\
\texttt{medium\_\allowbreak{}triangular\_\allowbreak{}opening\_\allowbreak{}pair\_\allowbreak{}03\_\allowbreak{}center\_\allowbreak{}x} & 30.11 & mm & N\\
\texttt{medium\_\allowbreak{}triangular\_\allowbreak{}opening\_\allowbreak{}pair\_\allowbreak{}03\_\allowbreak{}center\_\allowbreak{}y} & $-$7.95 & mm & N\\
\texttt{medium\_\allowbreak{}triangular\_\allowbreak{}opening\_\allowbreak{}pair\_\allowbreak{}03\_\allowbreak{}envelope\_\allowbreak{}width} & 36.23 & mm & N\\
\texttt{medium\_\allowbreak{}triangular\_\allowbreak{}opening\_\allowbreak{}pair\_\allowbreak{}03\_\allowbreak{}envelope\_\allowbreak{}height} & 66.89 & mm & N\\
\texttt{right\_\allowbreak{}triangular\_\allowbreak{}opening\_\allowbreak{}pair\_\allowbreak{}base\_\allowbreak{}length} & 73.30 & mm & N\\
\texttt{right\_\allowbreak{}triangular\_\allowbreak{}opening\_\allowbreak{}pair\_\allowbreak{}base\_\allowbreak{}angle} & 61.1613725 & deg & N\\
\texttt{right\_\allowbreak{}triangular\_\allowbreak{}opening\_\allowbreak{}01\_\allowbreak{}centroid\_\allowbreak{}x} & 55.50 & mm & N\\
\texttt{right\_\allowbreak{}triangular\_\allowbreak{}opening\_\allowbreak{}01\_\allowbreak{}centroid\_\allowbreak{}y} & $-$40.31 & mm & N\\
\texttt{central\_\allowbreak{}trapezoidal\_\allowbreak{}opening\_\allowbreak{}center\_\allowbreak{}y} & 96.50 & mm & N\\
\texttt{central\_\allowbreak{}trapezoidal\_\allowbreak{}opening\_\allowbreak{}corner\_\allowbreak{}radius} & 10.00 & mm & W\\
\texttt{central\_\allowbreak{}trapezoidal\_\allowbreak{}opening\_\allowbreak{}envelope\_\allowbreak{}width} & 184.30 & mm & W\\
\texttt{central\_\allowbreak{}trapezoidal\_\allowbreak{}opening\_\allowbreak{}envelope\_\allowbreak{}height} & 118.00 & mm & W\\
\texttt{central\_\allowbreak{}trapezoidal\_\allowbreak{}opening\_\allowbreak{}lower\_\allowbreak{}flat\_\allowbreak{}width} & 108.61 & mm & N\\
\texttt{small\_\allowbreak{}triangular\_\allowbreak{}opening\_\allowbreak{}centroid\_\allowbreak{}x} & 0.00 & mm & N\\
\texttt{small\_\allowbreak{}triangular\_\allowbreak{}opening\_\allowbreak{}centroid\_\allowbreak{}y} & 186.39 & mm & N\\
\texttt{small\_\allowbreak{}triangular\_\allowbreak{}opening\_\allowbreak{}base\_\allowbreak{}length} & 86.69 & mm & N\\
\texttt{small\_\allowbreak{}triangular\_\allowbreak{}opening\_\allowbreak{}base\_\allowbreak{}angle} & 43.8766972 & deg & N\\
\texttt{chamfer\_\allowbreak{}cut\_\allowbreak{}right\_\allowbreak{}triangular\_\allowbreak{}opening\_\allowbreak{}01\_\allowbreak{}center\_\allowbreak{}x} & $-$32.55 & mm & N\\
\texttt{chamfer\_\allowbreak{}cut\_\allowbreak{}right\_\allowbreak{}triangular\_\allowbreak{}opening\_\allowbreak{}01\_\allowbreak{}center\_\allowbreak{}y} & 203.69 & mm & N\\
\texttt{chamfer\_\allowbreak{}cut\_\allowbreak{}right\_\allowbreak{}triangular\_\allowbreak{}opening\_\allowbreak{}01\_\allowbreak{}envelope\_\allowbreak{}width} & 38.90 & mm & N\\
\texttt{chamfer\_\allowbreak{}cut\_\allowbreak{}right\_\allowbreak{}triangular\_\allowbreak{}opening\_\allowbreak{}01\_\allowbreak{}envelope\_\allowbreak{}height} & 37.63 & mm & N\\
\texttt{chamfer\_\allowbreak{}cut\_\allowbreak{}right\_\allowbreak{}triangular\_\allowbreak{}opening\_\allowbreak{}02\_\allowbreak{}center\_\allowbreak{}x} & 32.55 & mm & N\\
\texttt{chamfer\_\allowbreak{}cut\_\allowbreak{}right\_\allowbreak{}triangular\_\allowbreak{}opening\_\allowbreak{}02\_\allowbreak{}center\_\allowbreak{}y} & 203.69 & mm & N\\
\texttt{chamfer\_\allowbreak{}cut\_\allowbreak{}right\_\allowbreak{}triangular\_\allowbreak{}opening\_\allowbreak{}02\_\allowbreak{}envelope\_\allowbreak{}width} & 38.90 & mm & N\\
\texttt{chamfer\_\allowbreak{}cut\_\allowbreak{}right\_\allowbreak{}triangular\_\allowbreak{}opening\_\allowbreak{}02\_\allowbreak{}envelope\_\allowbreak{}height} & 37.63 & mm & N\\
\multicolumn{4}{@{}l}{\textit{Second right-triangular opening}}\\
\texttt{right\_\allowbreak{}triangular\_\allowbreak{}opening\_\allowbreak{}02\_\allowbreak{}centroid\_\allowbreak{}x} & $-$55.50 & mm & N\\
\texttt{right\_\allowbreak{}triangular\_\allowbreak{}opening\_\allowbreak{}02\_\allowbreak{}centroid\_\allowbreak{}y} & $-$40.31 & mm & N\\
\end{longtable}

}

\subsection{Asymmetric clevis joint connector (\code{freecad-fa0a1abef3})}
The key parameters are listed in Table~\ref{tab:example-parameters-fa0a1abef3}.

\begin{quote}
\small
An asymmetric clevis joint connector has a centered rectangular body, a large split-ring clevis at its positive-X end, a smaller fork at the negative-X end, and a long rounded side arm extending beyond the large clevis. Use global X along the connector, global Y across its width, and global Z through its thickness; the two ends are not mirror images.

The central body uses the listed connector\_body dimensions. The large clevis is centered at \texttt{clevis\_\allowbreak{}arc\_\allowbreak{}center\_\allowbreak{}x\_\allowbreak{}offset}, opens toward positive X, and uses \texttt{clevis\_\allowbreak{}end\_\allowbreak{}outer\_\allowbreak{}diameter}, \texttt{pin\_\allowbreak{}hole\_\allowbreak{}diameter}, \texttt{clevis\_\allowbreak{}slot\_\allowbreak{}width\_\allowbreak{}gap}, and the first-clevis tangent controls for its split-ring profile. Its inner and outer circular boundaries share a center, and the pin-hole cut enters from the positive-Z face. Its Z placement is set by \texttt{clevis\_\allowbreak{}body\_\allowbreak{}depth} and \texttt{clevis\_\allowbreak{}body\_\allowbreak{}depth\_\allowbreak{}offset}, and its pin-hole cut uses \texttt{pin\_\allowbreak{}hole\_\allowbreak{}center\_\allowbreak{}offset\_\allowbreak{}x} and \texttt{pin\_\allowbreak{}hole\_\allowbreak{}depth}. The \texttt{clevis\_\allowbreak{}body\_\allowbreak{}depth\_\allowbreak{}offset} is measured from the central body's positive-Z face.

The smaller fork opens toward negative X, with its rounded center one \texttt{clevis\_\allowbreak{}fork\_\allowbreak{}prong\_\allowbreak{}straight\_\allowbreak{}length} inward from the body end. Its concentric prong arcs, U-shaped opening, and positive-Z projection use the corresponding clevis\_fork and clevis\_slot parameters. The U-shaped opening is concentric with the rounded fork end and is cut from the positive-Z face toward negative Z. \texttt{clevis\_\allowbreak{}fork\_\allowbreak{}prong\_\allowbreak{}thickness} is the extrusion thickness rather than the radial wall thickness. This fork has no separate pin hole.

The long side arm lies against the front Y face and runs from \texttt{clevis\_\allowbreak{}arm\_\allowbreak{}profile\_\allowbreak{}start\_\allowbreak{}x} to the rounded end located by \texttt{clevis\_\allowbreak{}arm\_\allowbreak{}end\_\allowbreak{}center\_\allowbreak{}x} and \texttt{clevis\_\allowbreak{}arm\_\allowbreak{}end\_\allowbreak{}center\_\allowbreak{}z\_\allowbreak{}offset}. Its profile uses the listed clevis\_arm and clevis\_end dimensions, and \texttt{clevis\_\allowbreak{}pin\_\allowbreak{}hole\_\allowbreak{}diameter} defines a concentric hole through its rounded end parallel to global Y. The central, upper, and lower relief slots all belong to this arm: they begin at \texttt{clevis\_\allowbreak{}arm\_\allowbreak{}profile\_\allowbreak{}start\_\allowbreak{}x}, extend toward and open through the rounded end, and use their listed lengths, widths, transverse positions, and common \texttt{slot\_\allowbreak{}depth}. All three relief slots enter from the positive-Z face and cut toward negative Z.

\end{quote}

{\small
\setlength{\tabcolsep}{4pt}
\setlength{\LTcapwidth}{\linewidth}
\begin{longtable}{@{}l r l c@{}}
\caption{Key parameters of \texttt{freecad-fa0a1abef3}, as given to the agent. Check: W, a witness on the final geometry; C, a custom binding; N, a native sketch or feature measurement by the task's \texttt{param\_check.py}.}\label{tab:example-parameters-fa0a1abef3}\\
\toprule Parameter & Value & Unit & Check\\ \midrule \endfirsthead
\toprule Parameter & Value & Unit & Check\\ \midrule \endhead
\bottomrule \endlastfoot
\multicolumn{4}{@{}l}{\textit{Central body}}\\
\texttt{connector\_\allowbreak{}body\_\allowbreak{}length} & 75.0 & mm & N\\
\texttt{connector\_\allowbreak{}body\_\allowbreak{}width} & 45.0 & mm & N\\
\texttt{connector\_\allowbreak{}body\_\allowbreak{}thickness} & 12.0 & mm & W\\
\multicolumn{4}{@{}l}{\textit{Large split-ring clevis}}\\
\texttt{clevis\_\allowbreak{}end\_\allowbreak{}outer\_\allowbreak{}diameter} & 45.0 & mm & W\\
\texttt{pin\_\allowbreak{}hole\_\allowbreak{}diameter} & 25.0 & mm & W\\
\texttt{clevis\_\allowbreak{}slot\_\allowbreak{}width\_\allowbreak{}gap} & 5.0 & mm & W\\
\texttt{clevis\_\allowbreak{}arc\_\allowbreak{}center\_\allowbreak{}x\_\allowbreak{}offset} & 37.5 & mm & N\\
\texttt{clevis\_\allowbreak{}body\_\allowbreak{}depth} & 50.0 & mm & W\\
\texttt{clevis\_\allowbreak{}body\_\allowbreak{}depth\_\allowbreak{}offset} & $-$20.0 & mm & N\\
\texttt{first\_\allowbreak{}clevis\_\allowbreak{}outer\_\allowbreak{}slot\_\allowbreak{}tangent\_\allowbreak{}angle} & 6.3793702 & deg & N\\
\texttt{first\_\allowbreak{}clevis\_\allowbreak{}inner\_\allowbreak{}slot\_\allowbreak{}tangent\_\allowbreak{}angle} & 11.536959 & deg & N\\
\texttt{first\_\allowbreak{}clevis\_\allowbreak{}inner\_\allowbreak{}prong\_\allowbreak{}tangent\_\allowbreak{}x} & 49.7474487 & mm & N\\
\texttt{first\_\allowbreak{}clevis\_\allowbreak{}outer\_\allowbreak{}prong\_\allowbreak{}tangent\_\allowbreak{}x} & 59.8606798 & mm & N\\
\multicolumn{4}{@{}l}{\textit{Clevis pin-hole cut}}\\
\texttt{pin\_\allowbreak{}hole\_\allowbreak{}depth} & 34.5 & mm & N\\
\texttt{pin\_\allowbreak{}hole\_\allowbreak{}center\_\allowbreak{}offset\_\allowbreak{}x} & 37.5 & mm & N\\
\multicolumn{4}{@{}l}{\textit{Fork prongs}}\\
\texttt{clevis\_\allowbreak{}fork\_\allowbreak{}outer\_\allowbreak{}arc\_\allowbreak{}radius} & 17.5 & mm & W\\
\texttt{clevis\_\allowbreak{}fork\_\allowbreak{}inner\_\allowbreak{}arc\_\allowbreak{}radius} & 7.5 & mm & N\\
\texttt{clevis\_\allowbreak{}fork\_\allowbreak{}arc\_\allowbreak{}central\_\allowbreak{}angle} & 180 & deg & N\\
\texttt{clevis\_\allowbreak{}fork\_\allowbreak{}prong\_\allowbreak{}straight\_\allowbreak{}length} & 15.0 & mm & N\\
\texttt{clevis\_\allowbreak{}fork\_\allowbreak{}prong\_\allowbreak{}thickness} & 8.0 & mm & N\\
\multicolumn{4}{@{}l}{\textit{Fork U-shaped opening}}\\
\texttt{clevis\_\allowbreak{}slot\_\allowbreak{}arc\_\allowbreak{}radius} & 7.5 & mm & N\\
\texttt{clevis\_\allowbreak{}slot\_\allowbreak{}arc\_\allowbreak{}central\_\allowbreak{}angle} & 180.0 & deg & N\\
\texttt{clevis\_\allowbreak{}slot\_\allowbreak{}straight\_\allowbreak{}length} & 15.0 & mm & N\\
\texttt{clevis\_\allowbreak{}slot\_\allowbreak{}profile\_\allowbreak{}width} & 15.0 & mm & N\\
\texttt{clevis\_\allowbreak{}slot\_\allowbreak{}depth} & 34.5 & mm & N\\
\multicolumn{4}{@{}l}{\textit{Side arm}}\\
\texttt{clevis\_\allowbreak{}pin\_\allowbreak{}hole\_\allowbreak{}diameter} & 15.0 & mm & N\\
\texttt{clevis\_\allowbreak{}end\_\allowbreak{}arc\_\allowbreak{}radius} & 20.0 & mm & N\\
\texttt{clevis\_\allowbreak{}end\_\allowbreak{}arc\_\allowbreak{}central\_\allowbreak{}angle} & 180.0 & deg & N\\
\texttt{clevis\_\allowbreak{}arm\_\allowbreak{}length} & 46.7196386 & mm & N\\
\texttt{clevis\_\allowbreak{}arm\_\allowbreak{}width} & 40.0 & mm & N\\
\texttt{clevis\_\allowbreak{}arm\_\allowbreak{}depth} & 45.0 & mm & N\\
\texttt{clevis\_\allowbreak{}arm\_\allowbreak{}profile\_\allowbreak{}start\_\allowbreak{}x} & 52.9513389 & mm & N\\
\texttt{clevis\_\allowbreak{}arm\_\allowbreak{}end\_\allowbreak{}center\_\allowbreak{}x} & 99.6709775 & mm & N\\
\texttt{clevis\_\allowbreak{}arm\_\allowbreak{}end\_\allowbreak{}center\_\allowbreak{}z\_\allowbreak{}offset} & 9.0 & mm & N\\
\multicolumn{4}{@{}l}{\textit{Central relief slot}}\\
\texttt{clevis\_\allowbreak{}slot\_\allowbreak{}length} & 74.88 & mm & N\\
\texttt{clevis\_\allowbreak{}slot\_\allowbreak{}width} & 5.21 & mm & W\\
\texttt{clevis\_\allowbreak{}slot\_\allowbreak{}y\_\allowbreak{}min} & $-$2.71 & mm & N\\
\multicolumn{4}{@{}l}{\textit{Upper and lower relief slots}}\\
\texttt{upper\_\allowbreak{}slot\_\allowbreak{}length} & 72.25 & mm & N\\
\texttt{upper\_\allowbreak{}slot\_\allowbreak{}width} & 6.96 & mm & N\\
\texttt{upper\_\allowbreak{}slot\_\allowbreak{}y\_\allowbreak{}min} & 16.3556146 & mm & N\\
\texttt{lower\_\allowbreak{}slot\_\allowbreak{}length} & 73.8 & mm & N\\
\texttt{lower\_\allowbreak{}slot\_\allowbreak{}width} & 10.06 & mm & N\\
\texttt{lower\_\allowbreak{}slot\_\allowbreak{}y\_\allowbreak{}min} & $-$26.4156146 & mm & N\\
\texttt{slot\_\allowbreak{}depth} & 73.6 & mm & N\\
\end{longtable}

}

\section{Text-to-CAD Create-and-Edit Example Specification}
\label{app:example-edit-spec}
For \code{freecad-ff7883ab0c}, the following base description, edit instruction,
and parameters reproduce the supplied task specification. The shared output
contract requires a FreeCAD 1.1.0 script at \code{/app/answer.py} and two
documents derived from the script filename at runtime:
\code{answer_base.FCStd} and \code{answer_edit.FCStd}. Each must contain an
editable parametric feature tree in a single PartDesign Body producing one solid.

{\small

\paragraph{Base part.} An axisymmetric flanged inlet housing has a continuous central bore, a tapered inlet rim, a tubular middle body, a tapered rear hub, and a plain cylindrical rear flange disk. Along the axis, the rear hub widens into the flange disk, the body narrows through a short taper and continues cylindrically, and the inlet end expands through its tapered rim. No contoured front plate or rear-face circular opening pattern is present yet.

\paragraph{Edit instruction.} Add the front flange plate and rear opening pattern to the axisymmetric inlet body. Starting at \texttt{tapered\_\allowbreak{}inlet\_\allowbreak{}rim\_\allowbreak{}axial\_\allowbreak{}end\_\allowbreak{}position}, extend the fourfold-symmetric front plate forward by \texttt{square\_\allowbreak{}flange\_\allowbreak{}thickness} while preserving the existing central bore. Form its closed outline from the named side, corner, and transition arcs joined by the short tangent connectors. Add \texttt{number\_\allowbreak{}of\_\allowbreak{}bolt\_\allowbreak{}holes} equal through holes from the signed seed center and distribute them uniformly around the housing axis. From the rear cylindrical flange face, cut \texttt{number\_\allowbreak{}of\_\allowbreak{}flange\_\allowbreak{}base\_\allowbreak{}counterbores} equal circular openings with \texttt{flange\_\allowbreak{}base\_\allowbreak{}counterbore\_\allowbreak{}diameter} and \texttt{flange\_\allowbreak{}base\_\allowbreak{}counterbore\_\allowbreak{}depth}, again using the signed seed center and uniform angular spacing. Keep the tubular body, inlet rim, rear hub, rear disk envelope, and central passage unchanged.

}

{\small
\setlength{\tabcolsep}{4pt}
\setlength{\LTcapwidth}{\linewidth}
\begin{longtable}{@{}p{0.72\linewidth} r l@{}}
\caption{Text-to-CAD create-and-edit example parameters, reproduced from the agent instruction. The 15 base parameters remain unchanged; the edit adds the 14 target requirements below.}\label{tab:example-edit-parameters}\\
\toprule Parameter & Value & Unit\\ \midrule \endfirsthead
\toprule Parameter & Value & Unit\\ \midrule \endhead
\bottomrule \endlastfoot
\multicolumn{3}{@{}l}{\textit{Base part (unchanged by the edit)}}\\
\texttt{central\_\allowbreak{}bore\_\allowbreak{}diameter} & 50.8 & mm\\
\texttt{tubular\_\allowbreak{}body\_\allowbreak{}outer\_\allowbreak{}diameter} & 68.58 & mm\\
\texttt{tubular\_\allowbreak{}body\_\allowbreak{}taper\_\allowbreak{}end\_\allowbreak{}diameter} & 75.2 & mm\\
\texttt{tapered\_\allowbreak{}inlet\_\allowbreak{}rim\_\allowbreak{}small\_\allowbreak{}diameter} & 88.84 & mm\\
\texttt{tapered\_\allowbreak{}inlet\_\allowbreak{}rim\_\allowbreak{}large\_\allowbreak{}diameter} & 95.25 & mm\\
\texttt{tapered\_\allowbreak{}hub\_\allowbreak{}small\_\allowbreak{}diameter} & 92.08 & mm\\
\texttt{tapered\_\allowbreak{}hub\_\allowbreak{}large\_\allowbreak{}diameter} & 97.78 & mm\\
\texttt{rounded\_\allowbreak{}flange\_\allowbreak{}outer\_\allowbreak{}diameter} & 152.4 & mm\\
\texttt{tapered\_\allowbreak{}hub\_\allowbreak{}axial\_\allowbreak{}start\_\allowbreak{}position} & $-$27.36 & mm\\
\texttt{rounded\_\allowbreak{}flange\_\allowbreak{}axial\_\allowbreak{}start\_\allowbreak{}position} & $-$24.51 & mm\\
\texttt{tubular\_\allowbreak{}body\_\allowbreak{}axial\_\allowbreak{}start\_\allowbreak{}position} & $-$8.0 & mm\\
\texttt{tubular\_\allowbreak{}body\_\allowbreak{}axial\_\allowbreak{}middle\_\allowbreak{}position} & 7.57 & mm\\
\texttt{tubular\_\allowbreak{}body\_\allowbreak{}axial\_\allowbreak{}end\_\allowbreak{}position} & 26.8 & mm\\
\texttt{tapered\_\allowbreak{}inlet\_\allowbreak{}rim\_\allowbreak{}axial\_\allowbreak{}start\_\allowbreak{}position} & 26.8 & mm\\
\texttt{tapered\_\allowbreak{}inlet\_\allowbreak{}rim\_\allowbreak{}axial\_\allowbreak{}end\_\allowbreak{}position} & 30.0 & mm\\
\multicolumn{3}{@{}l}{\textit{Additional target requirements}}\\
\texttt{square\_\allowbreak{}flange\_\allowbreak{}thickness} & 16.51 & mm\\
\texttt{square\_\allowbreak{}flange\_\allowbreak{}side\_\allowbreak{}arc\_\allowbreak{}radius} & 63.5 & mm\\
\texttt{square\_\allowbreak{}flange\_\allowbreak{}corner\_\allowbreak{}arc\_\allowbreak{}radius} & 19.4447813 & mm\\
\texttt{square\_\allowbreak{}flange\_\allowbreak{}transition\_\allowbreak{}arc\_\allowbreak{}radius} & 25.4 & mm\\
\texttt{square\_\allowbreak{}flange\_\allowbreak{}corner\_\allowbreak{}arc\_\allowbreak{}center\_\allowbreak{}offset} & 40.132 & mm\\
\texttt{number\_\allowbreak{}of\_\allowbreak{}bolt\_\allowbreak{}holes} & 4 & \\
\texttt{bolt\_\allowbreak{}hole\_\allowbreak{}diameter} & 13.34 & mm\\
\texttt{bolt\_\allowbreak{}hole\_\allowbreak{}seed\_\allowbreak{}center\_\allowbreak{}x\_\allowbreak{}offset} & 42.6562166 & mm\\
\texttt{bolt\_\allowbreak{}hole\_\allowbreak{}seed\_\allowbreak{}center\_\allowbreak{}y\_\allowbreak{}offset} & 42.6562166 & mm\\
\texttt{flange\_\allowbreak{}base\_\allowbreak{}counterbore\_\allowbreak{}depth} & 34.54 & mm\\
\texttt{number\_\allowbreak{}of\_\allowbreak{}flange\_\allowbreak{}base\_\allowbreak{}counterbores} & 4 & \\
\texttt{flange\_\allowbreak{}base\_\allowbreak{}counterbore\_\allowbreak{}diameter} & 19.05 & mm\\
\texttt{flange\_\allowbreak{}base\_\allowbreak{}counterbore\_\allowbreak{}seed\_\allowbreak{}center\_\allowbreak{}x\_\allowbreak{}offset} & $-$42.72 & mm\\
\texttt{flange\_\allowbreak{}base\_\allowbreak{}counterbore\_\allowbreak{}seed\_\allowbreak{}center\_\allowbreak{}y\_\allowbreak{}offset} & $-$42.83 & mm\\
\end{longtable}

}

\section{Supplementary Reference-Model Statistics}
\begin{figure}[htbp]
\centering
\includegraphics[width=0.94\linewidth]{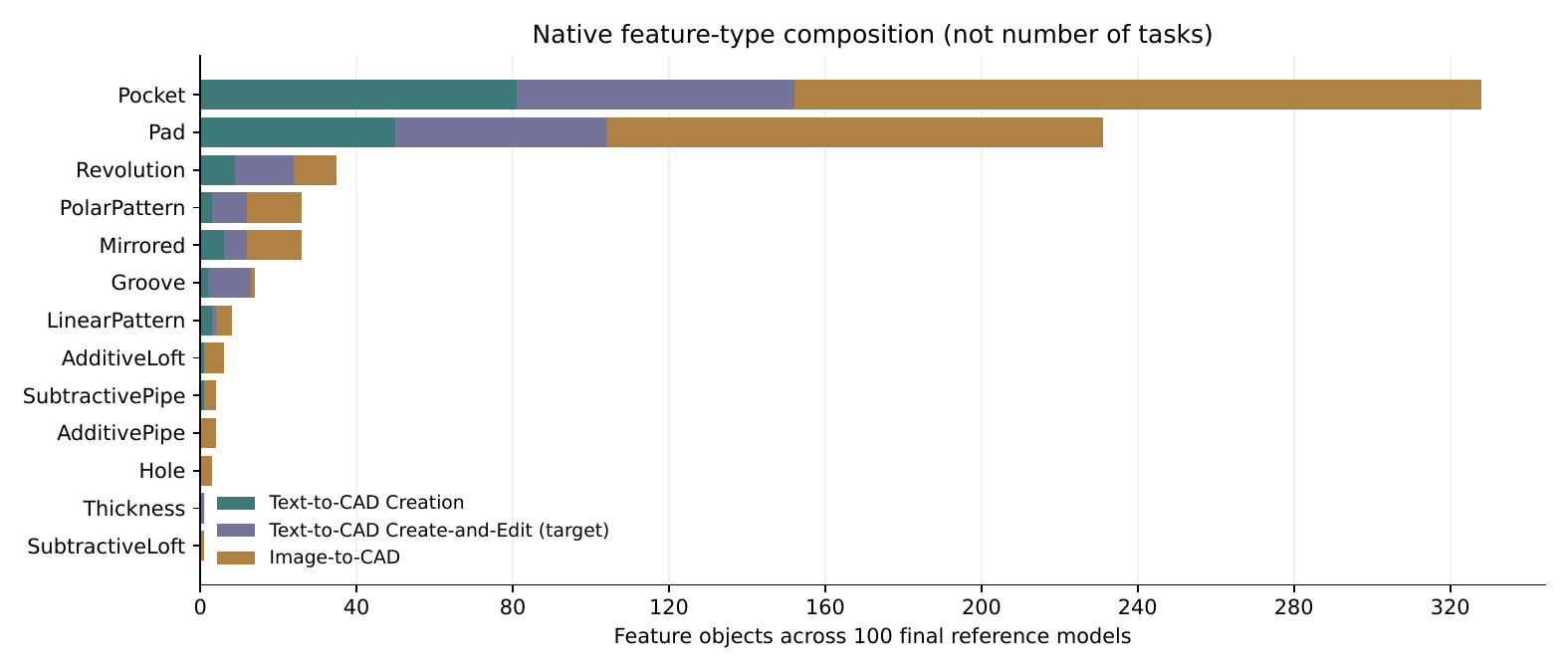}
\caption{Feature-object types across the 100 final references, broken down
by workflow. A task can contribute several objects of the same type.}
\label{fig:feature-types}
\end{figure}
\begin{figure}[htbp]
\centering
\includegraphics[width=\linewidth]{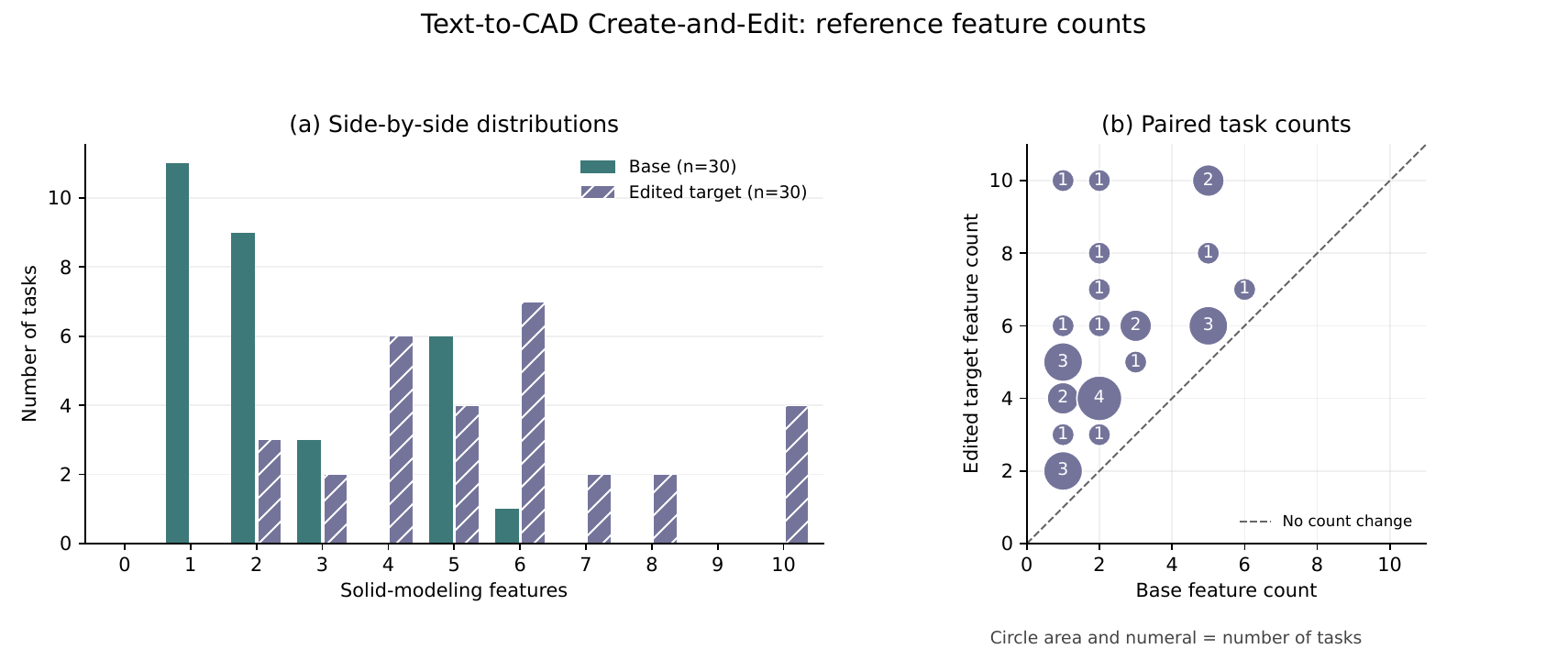}
\caption{Feature counts for the 30 text-to-CAD create-and-edit pairs. Left: side-by-side
bars distinguish base and edited target distributions. Right: paired base/target
counts; circle area and numerals show how many tasks share each pair. The dashed
diagonal indicates no count change. The target-minus-base count is not
necessarily the number of operations needed to perform an edit.}
\label{fig:edit-feature-counts}
\end{figure}
\begin{figure}[htbp]
\centering
\includegraphics[width=\linewidth]{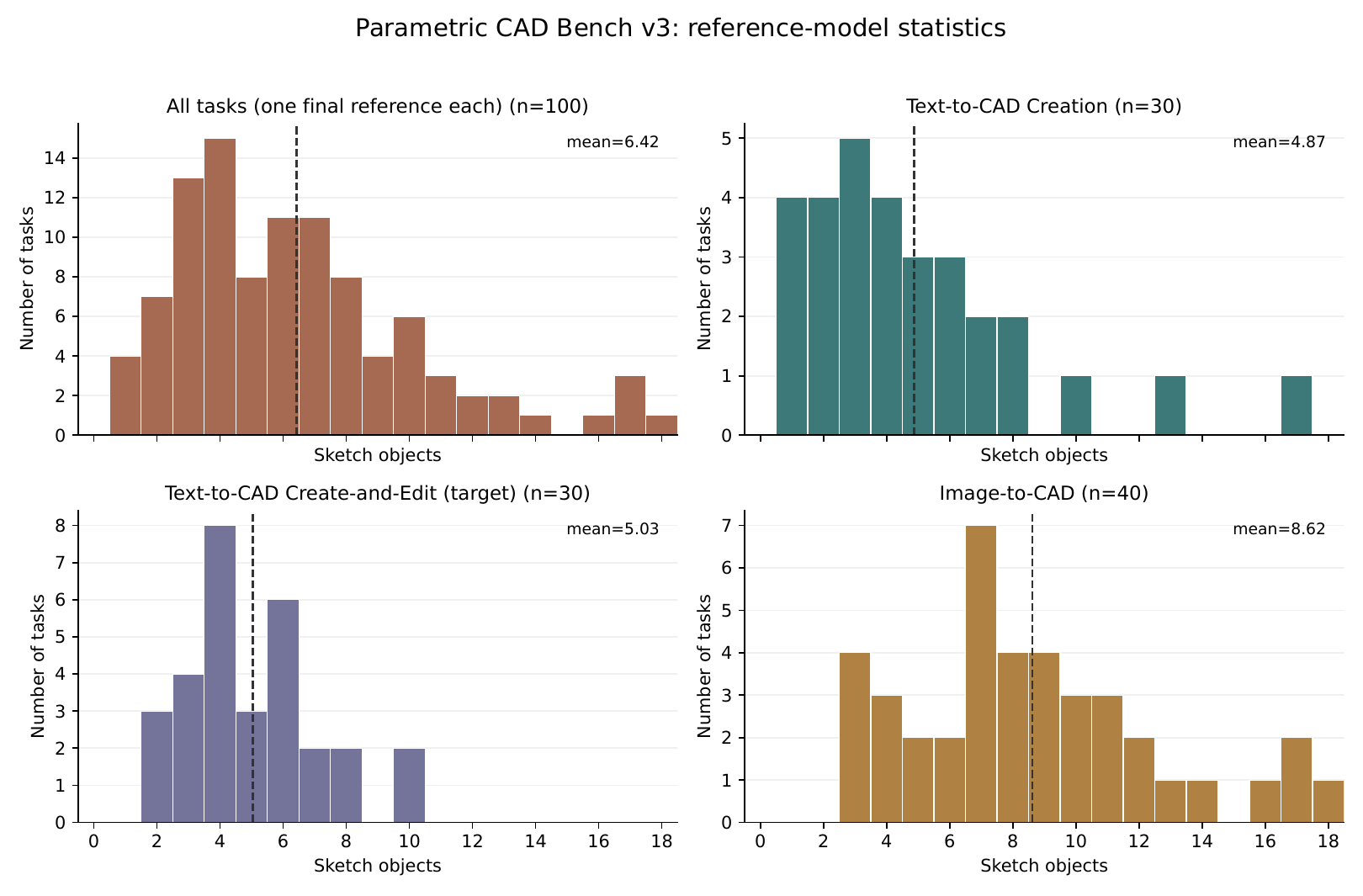}
\caption{Saved sketch-object counts per final reference model. Sketch objects
are excluded from the solid-modeling feature statistic.}
\label{fig:sketch-counts}
\end{figure}

\clearpage
\begin{landscape}
\section{Leaderboard Figures}
Each chart occupies a landscape page to keep all 13 model--agent labels
readable.
\begin{figure}[!h]
\centering
\includegraphics[width=0.96\linewidth,height=0.72\textheight,keepaspectratio]{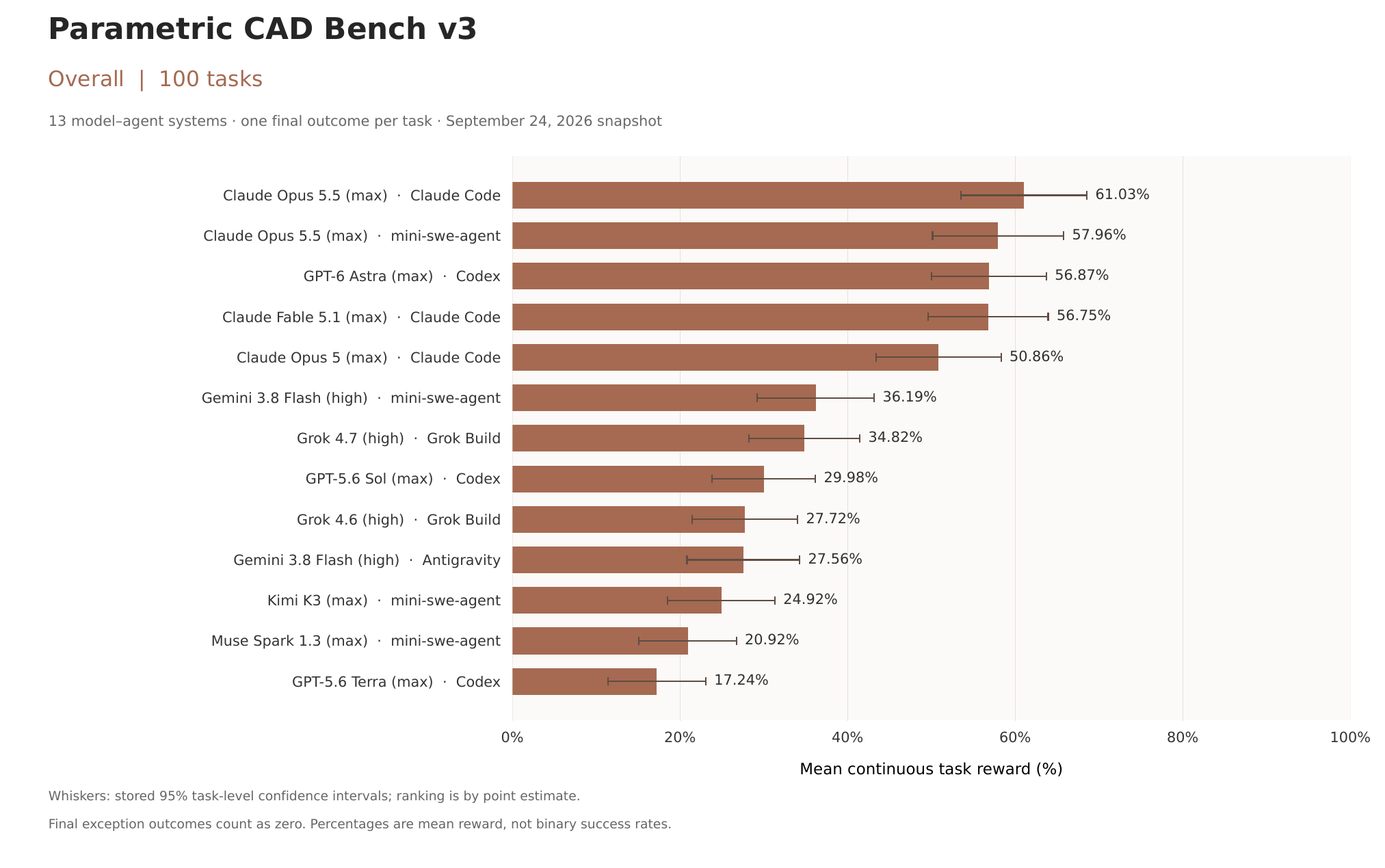}
\caption{Overall mean continuous reward across 100 tasks in the September 24, 2026
snapshot. Whiskers show 95\% task-level intervals. Every final
exception counts as zero. Ranking is by point estimate, not demonstrated
statistical separation.}
\label{fig:overall}
\end{figure}
\end{landscape}

\clearpage
\begin{landscape}
\begin{figure}[p]
\centering
\includegraphics[width=0.96\linewidth,height=0.82\textheight,keepaspectratio]{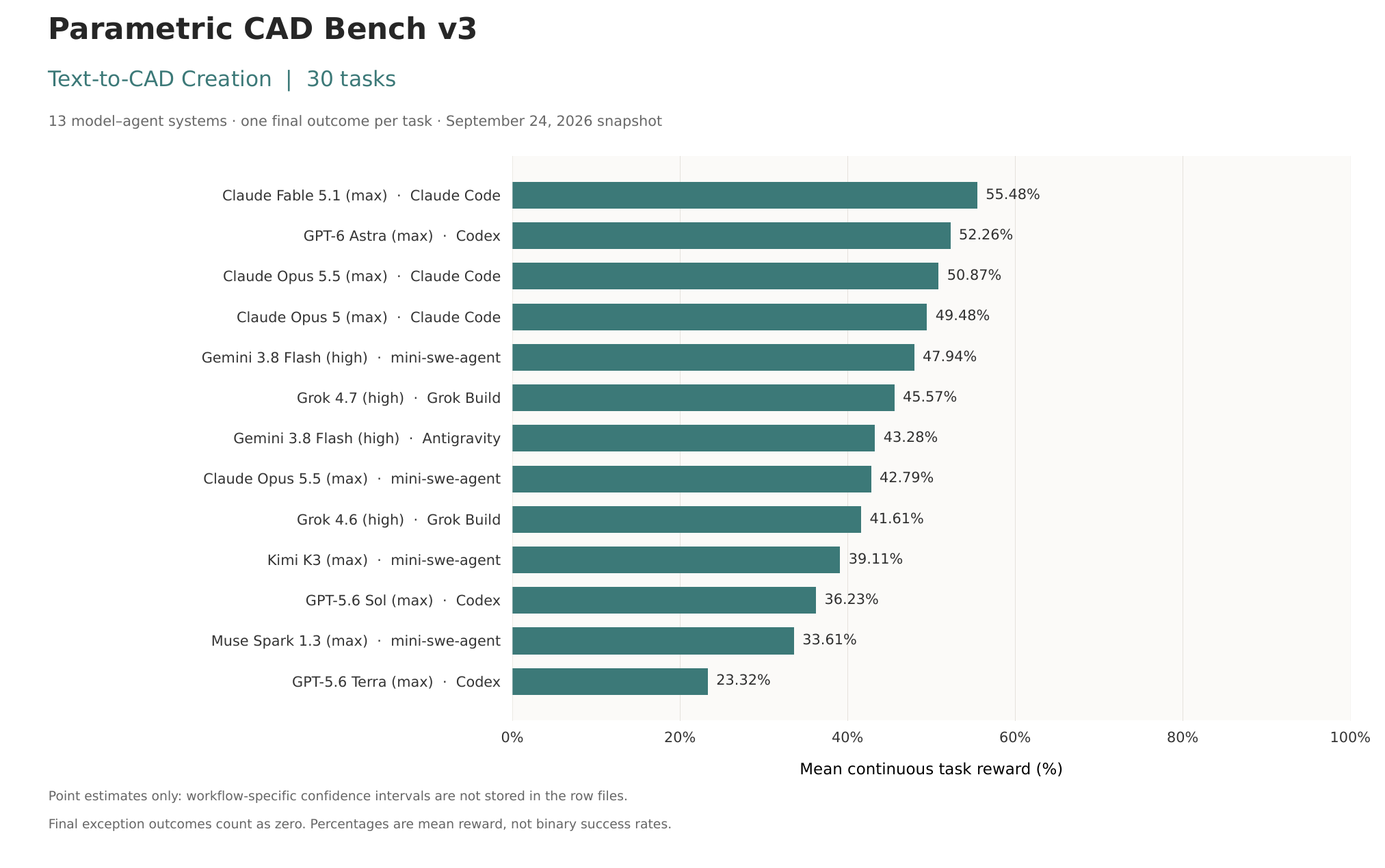}
\caption{Text-to-CAD creation mean reward across 30 tasks. Systems are ordered by their
creation point estimates; workflow-specific intervals are not plotted.}
\label{fig:create}
\end{figure}
\end{landscape}

\clearpage
\begin{landscape}
\begin{figure}[p]
\centering
\includegraphics[width=0.96\linewidth,height=0.82\textheight,keepaspectratio]{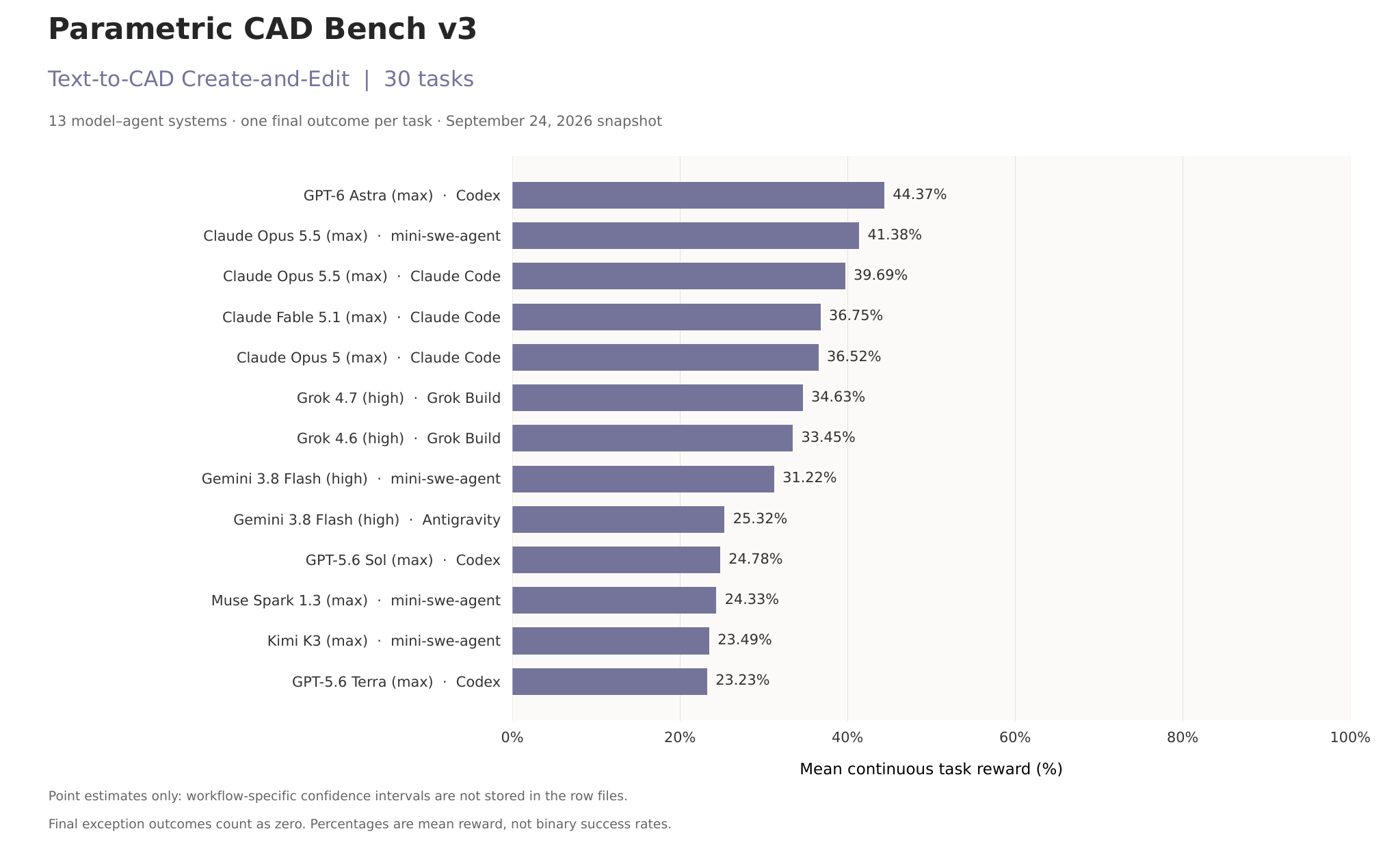}
\caption{Text-to-CAD create-and-edit mean reward across 30 tasks. Each task scores both
base and target artifacts through the nested harmonic aggregation in
Equation~\ref{eq:edit}. Bars show point estimates, not binary success rates.}
\label{fig:edit}
\end{figure}
\end{landscape}

\clearpage
\begin{landscape}
\begin{figure}[p]
\centering
\includegraphics[width=0.96\linewidth,height=0.82\textheight,keepaspectratio]{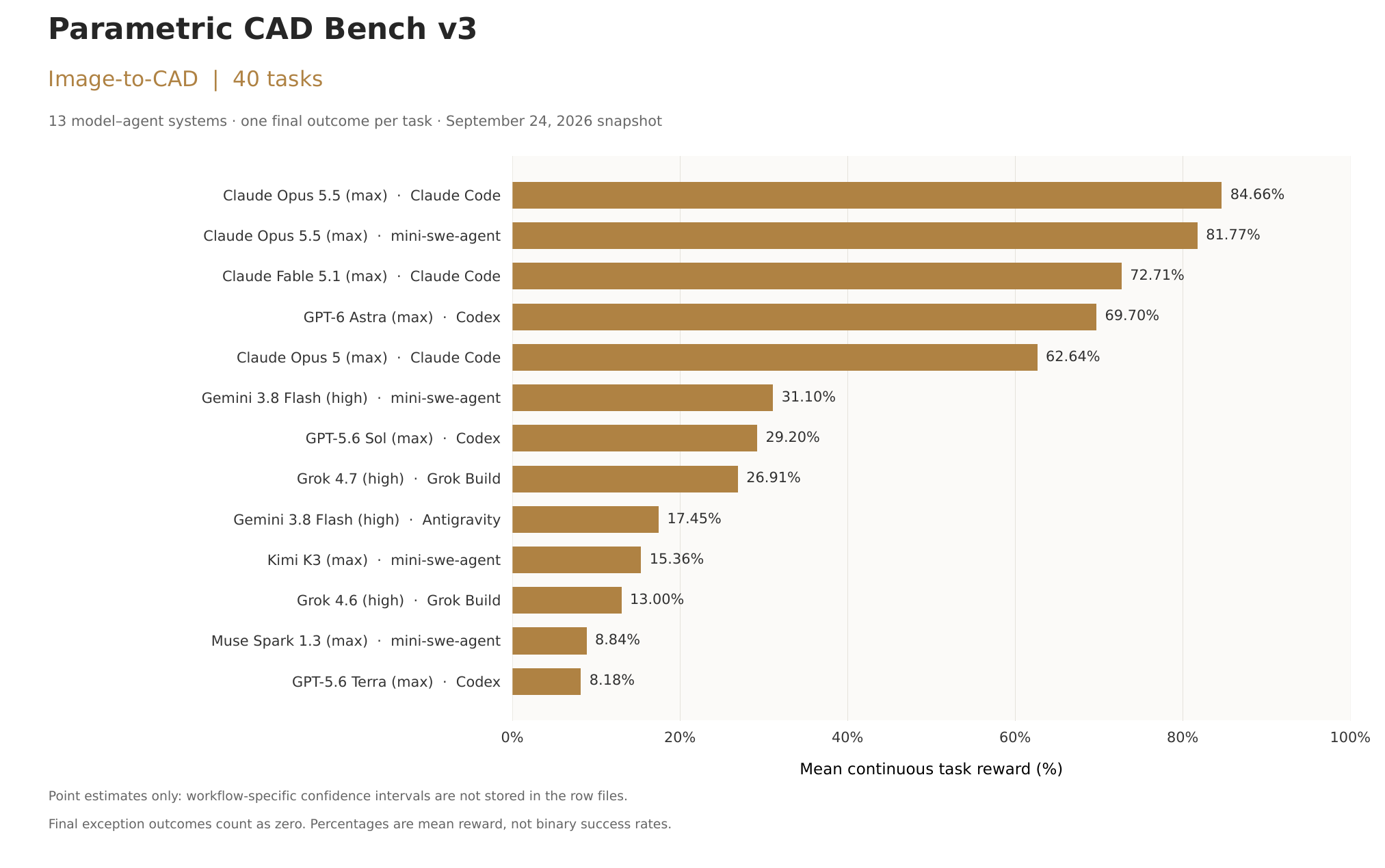}
\caption{Image-to-CAD mean reward across 40 engineering-drawing tasks. Bars show
workflow point estimates only. Different workflow populations prevent treating
cross-workflow differences as a controlled modality ablation.}
\label{fig:image}
\end{figure}
\end{landscape}

\end{document}